%% file: iclr2027_conference.tex
\documentclass{article} 
\PassOptionsToPackage{numbers,sort&compress}{natbib}
\usepackage{iclr2027_conference,times}

\input{math_commands.tex}

\usepackage{hyperref}
\usepackage{url}
\usepackage{booktabs}
\usepackage{multirow}
\usepackage{graphicx}
\usepackage[table]{xcolor}
\usepackage{algorithm}
\usepackage{algpseudocode}
\usepackage{mathtools}
\usepackage{amsmath}
\usepackage{amssymb}
\usepackage{enumitem}
\usepackage{bbm}

\hypersetup{hidelinks}

\algrenewcommand\algorithmicrequire{\textbf{Input:}}
\algrenewcommand\algorithmicensure{\textbf{Output:}}

\definecolor{bestred}{RGB}{200,0,0}
\definecolor{secondorange}{RGB}{220,120,0}

\title{What You Observe Determines How You 
\\ Identify Causal Effects: Evaluating Causal Models across Observational Views}

\author{
Heejin Jung$^{1}$ \quad
Gyeongdeok Seo$^{2}$ \quad
Hoyoon Byun$^{1}$ \quad
Joseph Lee$^{1}$ \quad
Kyungwoo Song$^{1}$\thanks{
Corresponding author: \texttt{kyungwoo.song@gmail.com}
} \\
$^{1}$Yonsei University
\qquad
$^{2}$University of Illinois Urbana--Champaign
}
\iclrfinalcopy 
\begin{document}
\raggedbottom
\setlength{\textfloatsep}{10pt}

\maketitle

\begin{abstract}
Causal foundation models (CFMs) pre-trained on data generated from various structural causal models (SCMs) have been proposed for
estimating causal effects from observational data. However, differences in pre-training environments and evaluation protocols make it difficult to assess how their performance depends on the information available for causal identification. To enable controlled comparisons, we introduce \textbf{CausalIDView}\footnote{Code is available at \url{https://github.com/MLAI-Yonsei/CausalIDView}.}, a multi-view benchmark that holds the SCM realization and target estimand fixed while varying only the observational view available to the estimator. Each observational view corresponds to a distinct identification regime under the benchmark's maintained causal assumptions. Across these matched views, no CFM consistently performs best and model rankings vary substantially. Under controlled structural changes, CFMs exhibit model-specific failures to maintain stable estimates when true effects are unchanged and to track genuine effect changes. We also examine whether combining explicit identification with strong predictive estimation is effective. A modular approach that pairs a predictive tabular foundation model with regime-specific identification procedures is competitive with CFMs and outperforms several of them. These findings motivate cross-regime comparisons to assess the empirical value of CFMs.
\end{abstract}

\vspace{-5pt}

\input{sec/1.Introduction}
\vspace{-5pt}
\input{sec/2.Background}
\vspace{-5pt}
\input{sec/3.Related_Work}
\vspace{-5pt}
\input{sec/4.Data_Curation}
\vspace{-5pt}
\input{sec/5.Experimental_Design}

\vspace{-5pt}
\input{sec/6.Result}
\vspace{-5pt}
\input{sec/7.Conclusion}

\clearpage
\section*{AI Use Statement}

We used large language models (LLMs) during manuscript preparation
to assist with language editing and to provide feedback on clarity,
logical flow, and presentation.
We also used LLMs for brainstorming and critical feedback on
experimental design and for discussing possible interpretations of
experimental results.
All research questions, methodological and experimental decisions,
analyses, and conclusions were determined and verified by the authors.
The authors reviewed and revised all AI-assisted content and take full
responsibility for the accuracy and integrity of the work.

\nocite{*}
\bibliographystyle{plainnat}
\bibliography{iclr2027_reference}

\newpage
\appendix

\section{Formal Problem Setup}
\label{app:formal_problem_setup}

This appendix clarifies the distinction between the generating causal
model, its realized data, and the information available to an estimator.
It also specifies the identification terminology and the scope of the
synthetic benchmark.

\subsection{Identification Terminology}
\label{app:identification_terminology}

\paragraph{Identified set for a general estimand.}
Let $O$ denote an observed random vector with population distribution
$P_{\mathrm{obs}}$, and let $\mathcal A$ denote a collection of
maintained causal assumptions. Let $\mathcal M(\mathcal A)$ denote the
class of full-data causal laws satisfying $\mathcal A$, and consider a
generic causal estimand
\[
    \theta(Q)\in\Theta,
\]
where $Q$ denotes a candidate full-data causal law and $\Theta$ is the
corresponding parameter space. Following the identified-set formulation
\citep{manski2003partial}, the identified set is
\begin{equation}
    \mathcal I_{\theta}
    \bigl(P_{\mathrm{obs}};\mathcal A\bigr)
    :=
    \left\{
        \theta(Q)\in\Theta:
        Q\in\mathcal M(\mathcal A),\;
        Q_O=P_{\mathrm{obs}}
    \right\},
    \label{eq:app_general_identified_set}
\end{equation}
where $Q_O$ denotes the observed-data distribution induced by $Q$.
Hence,
$\mathcal I_{\theta}(P_{\mathrm{obs}};\mathcal A)$ contains all values
of the target estimand that are compatible with both the observed-data
distribution and the maintained assumptions.

\paragraph{Point identification, partial identification,
and no identifying information.}
Let $Q_0$ denote the true full-data causal law and
$\theta_0:=\theta(Q_0)$. Assuming
$Q_0\in\mathcal M(\mathcal A)$, we distinguish
\begin{alignat*}{3}
    &\mathcal I_{\theta}(P_{\mathrm{obs}};\mathcal A)
    &&{}=\{\theta_0\},
    &\qquad&\text{point identification},
    \\
    \{\theta_0\}\subsetneq{}
    &\mathcal I_{\theta}(P_{\mathrm{obs}};\mathcal A)
    &&{}\subsetneq\Theta,
    &\qquad&\text{partial identification},
    \\
    &\mathcal I_{\theta}(P_{\mathrm{obs}};\mathcal A)
    &&{}=\Theta,
    &\qquad&\text{no identifying information}.
\end{alignat*}

\begin{figure*}[t]
    \centering
    \setlength{\abovecaptionskip}{2pt}
    \setlength{\parskip}{0pt}
    \includegraphics[width=\textwidth]{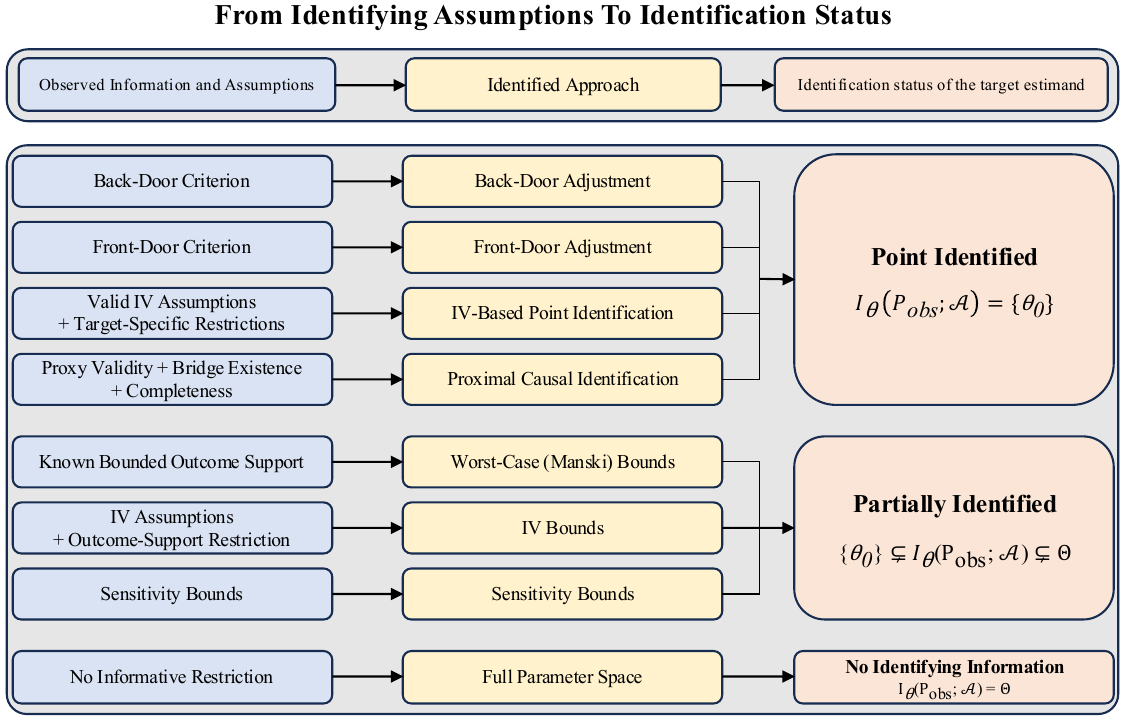}
    \caption{
    \textbf{Taxonomy of identification.}
    The three identification status categories are defined by the identified
    set. The assumptions and identification approaches shown are
    representative examples, including those relevant to our benchmark,
    rather than an exhaustive enumeration of possible identification regimes.
    }
    \label{fig:Taxonomy}
\end{figure*}

We use non-point identification for the broader case in which
the identified set is not a singleton. It therefore includes
informative partial identification and should not be equated with the
absence of any information about the target. The final case above is
the limiting situation in which the observed distribution and
maintained assumptions rule out no value in the original parameter
space. Identification status is consequently always relative to a
specified target, observed-data distribution, and set of maintained
assumptions, as summarized in
Figure~\ref{fig:Taxonomy}.

We now specialize the general definition to the target considered in
this work. For a fixed covariate value $x$, define under a candidate
causal law $Q$
\begin{equation*}
    \tau_Q(x)
    :=
    \mathbb E_Q\!\left[
        Y(1)-Y(0)\mid X=x
    \right].
\end{equation*}
Taking $\theta(Q)=\tau_Q(x)$
in Equation~\ref{eq:app_general_identified_set}, the identified set for the target CATE in regime r is therefore
\begin{equation*}
    \Theta_I^{(r)}(x)
    :=
    \mathcal I_{\tau(x)}
    \bigl(P_{\mathrm{obs}}^{(r)};\mathcal A_r\bigr)
    =
    \left\{
        \tau_Q(x)\in\Theta:
        Q\in\mathcal M(\mathcal A_r),\;
        Q_{O^{(r)}}=P_{\mathrm{obs}}^{(r)}
    \right\}.
\end{equation*}
The true target is
$\theta_0=\tau_0(x)$. Across the observational regimes in
\textsc{CausalIDView}, $\tau_0(x)$ is held fixed, while
$O^{(r)}$, $P_{\mathrm{obs}}^{(r)}$, and the assumptions
$\mathcal A_r$ may differ.

When $\Theta_I^{(r)}(x)=\{\tau_0(x)\}$,
the target is point identified at $x$. Equivalently, there exists a
regime-specific observed-data functional $\Phi_x^{(r)}$ such that $\tau_0(x)
    =
    \Phi_x^{(r)}
    \!\left(P_{\mathrm{obs}}^{(r)}\right)
    \text{under }\mathcal A_r.
    \label{eq:app_identifying_functional}$
Different observational views may identify the same target through different observed-data functionals. Conversely, the absence of a point-identification claim for the HC view does not by itself
imply
$\Theta_I^{(\mathrm{HC})}(x)=\Theta$.
The assumptions and identifying functionals associated with each
regime are detailed in Appendix~\ref{app:identification_regimes}.

\paragraph{Identification versus statistical estimation.}
Identification concerns whether the target is uniquely determined by
the population observed-data law under maintained causal assumptions
\citep{JMLR:v9:shpitser08a,shpitser2020identification}.
Statistical estimation instead concerns recovering an identified
functional, or an identified set, from finite data. A target may
therefore be point identified while remaining statistically difficult
to estimate. Conversely, a point prediction may happen to be close to
the oracle effect even when point identification has not been
established from the available observational view.

\subsection{Benchmark Scope}
\label{app:benchmark_scope}

The synthetic generator does not include longitudinal or time-varying
treatments, dynamic treatment regimes, interference between units,
continuous or multivalued treatments, censoring or survival outcomes,
or multidimensional latent confounders.

\section{Identification Regimes}
\label{app:identification_regimes}

We expand the identification arguments in Section~\ref{sec:background}
 and show how each observational view supports the same target CATE. Structural arguments are distinguished from finite-sample diagnostics,
which are reported separately in Appendix~\ref{sec:data-validation}.

\subsection{Back-Door}
\label{app:backdoor_identification}

\paragraph{Assumptions.}
The back-door view observes $(X,U,T,Y)$, with $(X,U)$ measured before
treatment. In addition to the common consistency conditions, we require:
\begin{enumerate}
    \setlength{\itemsep}{0pt}
    \item \textit{Conditional exchangeability:}
    $\{Y(0),Y(1)\}\perp T\mid X,U$.
    \item \textit{Positivity:}
    $0<P(T=1\mid X=x,U=u)<1$ for every $(x,u)$ in the target support.
\end{enumerate}

\paragraph{Recovering the CATE.}
Let $m_t(x,u):=\mathbb E[Y\mid T=t,X=x,U=u]$.
Exchangeability and consistency allow the potential-outcome mean to be
computed by averaging the observed outcome regression over the same
confounder distribution in both arms \citep{pearl2022causal}:
\begin{equation*}
    \mu_t^{\mathrm{BD}}(x)
    =\mathbb E[Y(t)\mid X=x]
    =\sum_{u\in\{-1,+1\}}m_t(x,u)P(U=u\mid X=x).
\end{equation*}
For the SCM outcome law,
$m_t(x,u)=\mu(x)+t\tau_0(x)+\gamma_U(x)u.$
The prognostic and confounding terms therefore cancel within each
$(x,u)$ stratum, giving
\begin{align*}
    \tau_{\mathrm{BD}}(x)
    &=\sum_u\{m_1(x,u)-m_0(x,u)\}P(U=u\mid X=x)\\
    &=\sum_u\tau_0(x)P(U=u\mid X=x)=\tau_0(x).
\end{align*}

\begin{table}[t]
    \centering
    \caption{
        Observational contexts constructed from each complete causal
        world. All views share the same context and query units,
        treatment assignments, factual outcomes, and oracle effects.
    }
    \label{tab:observational_views}
    \small
    \begin{tabular}{ll}
        \hline
        View & Observed context variables \\
        \hline
        Back-door (BD)
        & $X,U,T,Y$ \\
        Front-door (FD)
        & $X,M,T,Y$ \\
        Instrumental variable (IV)
        & $X,I,T,Y$ \\
        Proximal (PX)
        & $X,Z_{\mathrm{p}},W_{\mathrm{p}},T,Y$ \\
        Hidden Confounding (HC)
        & $X,T,Y$ \\
        \hline
    \end{tabular}
\end{table}

\subsection{Front-Door}
\label{app:frontdoor_identification}

\paragraph{Assumptions.}
The front-door view observes $(X,T,M,Y)$ and hides $U$.
Conditional on $X$, we require \citep{pearl2022causal}:
\begin{enumerate}
    \setlength{\itemsep}{0pt}
    \item \textit{Complete mediation:}
    $M$ intercepts every directed path from $T$ to $Y$.
    \item \textit{Treatment--mediator exchangeability:}
    no back-door path between $T$ and $M$ remains open given $X$.
    \item \textit{Mediator--outcome adjustment:}
    all back-door paths between $M$ and $Y$ are blocked by $(T,X)$.
    \item \textit{Positivity and support:}
    both treatment arms have positive probability given $X$, and
    mediator values used in the functional are supported in both arms.
\end{enumerate}

\paragraph{Recovering the CATE.}
Write $Q(m,s,x):=\mathbb E[Y\mid M=m,T=s,X=x]$ and
$p_s(x):=P(T=s\mid X=x)$.
In the SCM, $M\perp U\mid T,X$, so
\begin{equation*}
    Q(m,s,x)=\mu(x)+\beta(x)m
      +\gamma_U(x)\mathbb E[U\mid T=s,X=x].
\end{equation*}
First, average this regression over the observed treatment distribution:
\begin{align*}
    G(m,x)&:=\sum_{s\in\{0,1\}}Q(m,s,x)p_s(x)\\
    &=\mu(x)+\beta(x)m
      +\gamma_U(x)\sum_s\mathbb E[U\mid T=s,X=x]p_s(x)\\
    &=\mu(x)+\beta(x)m,
\end{align*}
where the last equality uses
$\sum_s\mathbb E[U\mid T=s,X=x]p_s(x)=\mathbb E[U\mid X=x]=0$.
This step removes the treatment-specific latent composition from the
outcome regression. Next, integrate over the mediator distribution
under each treatment arm:
\begin{align*}
    \mu_t^{\mathrm{FD}}(x)
      &=\int G(m,x)p(m\mid T=t,X=x)\,dm\\
      &=\mu(x)+\beta(x)\mathbb E[M\mid T=t,X=x]
       =\mu(x)+\beta(x)\delta t.
\end{align*}
Consequently,
\begin{equation*}
    \tau_{\mathrm{FD}}(x)
      =\mu_1^{\mathrm{FD}}(x)-\mu_0^{\mathrm{FD}}(x)
      =\beta(x)\delta=\tau_0(x).
\end{equation*}
The total effect thus requires integrating over the two mediator laws,
not holding the query unit's factual mediator fixed.

\subsection{Instrumental Variable}
\label{app:iv_identification}

\paragraph{Assumptions.}
The IV view observes $(X,I,T,Y)$.
Let $T^I(j)$ denote potential treatment under instrument value $j$.
We use the following conditions:
\begin{enumerate}
    \setlength{\itemsep}{0pt}
    \item \textit{Exogeneity:}
    $I\perp\{Y(0),Y(1),T^I(0),T^I(1)\}\mid X$.
    \item \textit{Exclusion:}
    $I$ affects $Y$ only through $T$.
    \item \textit{Instrument positivity and relevance:}
    $0<P(I=1\mid X=x)<1$ and
    $\mathbb E[T\mid I=1,X=x]\neq\mathbb E[T\mid I=0,X=x]$.
    \item \textit{Monotonicity:}
    $T^I(1)\geq T^I(0)$ almost surely.
    \item \textit{Within-$X$ gain invariance:}
    the benchmark additionally imposes $Y(1)-Y(0)=\tau_0(X)$,
    so gains do not vary with latent state or compliance type at fixed $X$.
\end{enumerate}

\paragraph{Recovering the CATE.}
The first four conditions give the conditional Wald ratio a complier-effect interpretation and the final restriction makes that effect equal to the
CATE \citep{angrist1996identification}.
To see this directly, the SCM has $I\perp U\mid X$ and
$\mathbb E[U\mid X=x]=0$, hence
\begin{align*}
    \mathbb E[Y\mid I=j,X=x]
    &=\mu(x)+\tau_0(x)\mathbb E[T\mid I=j,X=x]
      +\gamma_U(x)\mathbb E[U\mid I=j,X=x]\\
    &=\mu(x)+\tau_0(x)\mathbb E[T\mid I=j,X=x].
\end{align*}
Define the instrument-induced differences
\begin{equation*}
    \begin{aligned}
    \Delta_Y(x)&:=\mathbb E[Y\mid I=1,X=x]
                   -\mathbb E[Y\mid I=0,X=x],\\
    \Delta_T(x)&:=\mathbb E[T\mid I=1,X=x]
                   -\mathbb E[T\mid I=0,X=x].
    \end{aligned}
\end{equation*}
Subtracting the two reduced-form means cancels $\mu(x)$ and yields
\begin{equation*}
    \Delta_Y(x)=\tau_0(x)\Delta_T(x),
    \qquad
    \tau_{\mathrm{IV}}(x)=\frac{\Delta_Y(x)}{\Delta_T(x)}=\tau_0(x).
\end{equation*}
Thus, dividing the instrument-induced outcome change by the
instrument-induced treatment change recovers the shared CATE under
the additional gain restriction.

\subsection{Proximal Identification}
\label{app:proximal_identification}

\paragraph{Assumptions.}
The proximal view observes $(X,Z_p,W_p,T,Y)$, where $Z_p$ and $W_p$
are the treatment- and outcome-inducing proxies, respectively.
We require \citep{miao2018identifying,tchetgen2024introduction}:
\begin{enumerate}
    \setlength{\itemsep}{0pt}
    \item \textit{Latent exchangeability:}
    $Y(t)\perp T\mid U,X$.
    \item \textit{Proxy restrictions:}
    $Y\perp Z_p\mid T,U,X$ and
    $W_p\perp(T,Z_p)\mid U,X$.
    \item \textit{Bridge existence:}
    an integrable outcome bridge $h_t$ satisfies
    \begin{equation*}
        \mathbb E[Y\mid Z_p=z,T=t,X=x]
        =\mathbb E[h_t(W_p,x)\mid Z_p=z,T=t,X=x].
    \end{equation*}
    \item \textit{Completeness:}
    for $V=U$ and $V=W_p$, and every square-integrable $g$,
    \begin{equation*}
        \mathbb E[g(V)\mid Z_p,T=t,X=x]=0
        \quad\Longrightarrow\quad g(V)=0
    \end{equation*}
    almost surely under the corresponding conditional law.
    In this binary construction, these are full-rank conditions.
    \item \textit{Support:}
    both latent strata have positive probability given $X$,
    $0<P(T=1\mid X,U)<1$, and the required proxy cells have
    positive conditional probability.
\end{enumerate}

\paragraph{Recovering the CATE.}
For binary proxies, define the observed matrix and mean vector by
\begin{equation*}
    A_t(x)_{z,w}:=P(W_p=w\mid Z_p=z,T=t,X=x),
    \qquad
    y_t(x)_z:=\mathbb E[Y\mid Z_p=z,T=t,X=x]
\end{equation*}
for $z,w\in\{0,1\}$.
The bridge equation is a two-equation linear system,
\begin{equation*}
    A_t(x)\boldsymbol h_t(x)=\boldsymbol y_t(x),
    \qquad
    \boldsymbol h_t(x)=A_t(x)^{-1}\boldsymbol y_t(x),
\end{equation*}
where $\boldsymbol h_t(x)=(h_t(0,x),h_t(1,x))^\top$.
After solving this system for each arm, average the bridge values over
$W_p\mid X=x$:
\begin{equation*}
    \mu_t^{\mathrm{PX}}(x)
      =\sum_{w\in\{0,1\}}h_t(w,x)P(W_p=w\mid X=x).
\end{equation*}
For the SCM, write
$p_W^\pm(x):=P(W_p=1\mid U=\pm1,X=x)$.
An explicit bridge is
\begin{equation*}
    h_t(w,x)=\mu(x)+t\tau_0(x)-\gamma_U(x)
       +\frac{2\gamma_U(x)}{p_W^+(x)-p_W^-(x)}
          \{w-p_W^-(x)\}.
\end{equation*}
Only the term $t\tau_0(x)$ changes between treatment arms, so
$h_1(w,x)-h_0(w,x)=\tau_0(x)$. Therefore,
\begin{equation*}
    \tau_{\mathrm{PX}}(x)
      =\sum_w\{h_1(w,x)-h_0(w,x)\}P(W_p=w\mid X=x)
      =\tau_0(x).
\end{equation*}

\subsection{Hidden Confounding and Partial Identification}
\label{app:hidden_confounding}
The Hidden Confounding (HC) view observes only $(X,T,Y)$.
For the main continuous-outcome SCM, its outcome law and the zero
conditional mean of its noise give
\begin{equation*}
    \mathbb E[Y\mid T=t,X=x]
      =\mu(x)+t\tau_0(x)+\gamma_U(x)\mathbb E[U\mid T=t,X=x].
\end{equation*}
Subtracting the observed treatment-arm means therefore gives
\begin{equation}
    \begin{gathered}
    \underbrace{\mathbb E[Y\mid T=1,X=x]
         -\mathbb E[Y\mid T=0,X=x]}_{\text{observed group contrast}}\\[4pt]
    =\underbrace{\tau_0(x)}_{\text{target CATE}}
    +\underbrace{\gamma_U(x)\bigl\{
        \mathbb E[U\mid T=1,X=x]
        -\mathbb E[U\mid T=0,X=x]\bigr\}}
       _{\text{remaining confounding bias }b_{\mathrm{HC}}(x)}.
    \end{gathered}
    \label{eq:appB_hc_bias}
\end{equation}
The term $\gamma_U(X)U$ does not cancel when comparing different treatment groups, because treatment selection changes
their conditional distributions of $U$. Confounding bias remains
even though $U$ does not modify the treatment gain. Since this bias can
vary with $x$, it can distort the shape of the observational contrast
as well as its mean. Equation~\ref{eq:appB_hc_bias} describes the observational contrast,
not necessarily the output of every evaluated model.
We score HC point predictions against the oracle $\tau_0(x)$ as a
hidden-confounding stress test, without a point-identification claim.

We distinguish the three partial-identification approaches in
Figure~\ref{fig:Taxonomy}: Worst-Case (Manski) bounds, IV bounds,
and sensitivity bounds.
These evaluations use a separate partial-identification companion SCM described in Appendix~\ref{app:partial_id_scm}, distinct from the SCM used for the point-estimation benchmark.
The Manski and sensitivity views observe $(X,T,Y)$, whereas the IV view
additionally observes $I$.
The three views share the same realized $(X,T,Y)$,
context--query split, query units, and target CATE, enabling paired
comparisons across the partial-identification settings while varying
the available identifying information and assumptions.
Following Appendix~\ref{app:identification_terminology}, each identified
set is defined relative to its observed distribution and maintained
assumptions.

For finite-sample estimation, we keep the corresponding identifying
functional fixed and estimate the required observable quantities from
the context data.
The Manski and IV estimators first estimate the relevant cell
probabilities and then apply the closed-form Manski bounds or the
monotone-IV linear program, respectively.
If estimated IV cells are infeasible, they are projected onto the
monotone-IV response-type polytope before endpoint computation.
For sensitivity bounds, we evaluate CSA-PFN (MSM)~\citep{javurek2026amortizing}, B-Learner~\citep{oprescu2023b}, and
NeuralCSA~\citep{frauen2024neural} against a common numerical MSM reference.
Detailed implementations of the partial-identification estimators are
provided in Appendix~\ref{app:partial_id_baseline}.
Throughout, we distinguish the population identified set,
its numerical reference, and the estimated interval.
For estimator $m$, let
$[\widehat L_m(x),\widehat U_m(x)]$
denote its estimated interval.
Target-CATE containment denotes whether $\tau_0(x)$ lies in an estimated
set, rather than confidence-interval coverage.

\subsubsection{Worst-case (Manski) bounds}
For binary potential outcomes $Y(t)\in\{0,1\}$, define the four
observed cells $q_{ty}(x):=P(T=t,Y=y\mid X=x)$.
Under consistency alone, factual outcomes constrain the observed arm while missing
counterfactuals remain unrestricted within the binary support.
The resulting sharp CATE set is \citep{manski2003partial}
\begin{equation}
\begin{aligned}
    \Theta_I^{(\mathrm M)}(x)&=[L_{\mathrm M}(x),U_{\mathrm M}(x)],\\
    L_{\mathrm M}(x)&=-q_{10}(x)-q_{01}(x),\\
    U_{\mathrm M}(x)&= q_{11}(x)+q_{00}(x).
\end{aligned}
\label{eq:appB_manski_set}
\end{equation}
The width is exactly one because the four cells sum to one.
These bounds use the potential-outcome support $\{0,1\}$.

\subsubsection{Monotone-IV bounds}
Let $q_{jty}(x):=P(T=t,Y=y\mid I=j,X=x)$ for the matched IV view.
We maintain consistency, conditional instrument exogeneity and
exclusion, instrument positivity and relevance, and treatment
monotonicity $T^I(1)\geq T^I(0)$.
Unlike the point-identification argument in
Appendix~\ref{app:iv_identification}, this identified-set model does
not impose within-$X$ gain invariance or equality of mean effects
across compliance types.
The conditional response-type law is represented by 12 probabilities:
three treatment types crossed with all four binary outcome types,
\begin{equation*}
    r=(d_0^r,d_1^r,y_0^r,y_1^r)\in\mathcal R
    :=\{(0,0),(0,1),(1,1)\}\times\{0,1\}^2.
\end{equation*}
Here $d_j^r$ is the treatment under $I=j$ and $y_t^r$ is the
outcome under $T=t$. Let $q^{\mathrm{IV}}(x)$ stack the eight
conditional cells and define
\begin{equation*}
\begin{aligned}
    A_{(j,t,y),r}&:=\mathbf 1\{d_j^r=t,\ y_t^r=y\},\\
    \Pi_x&:=\{\pi\in\mathbb R^{12}:\ \pi\geq0,\
        \mathbf 1^\top\pi=1,\ A\pi=q^{\mathrm{IV}}(x)\}.
\end{aligned}
\end{equation*}
The same response-type probabilities reproduce both instrument arms;
this encodes conditional exogeneity, while the mapping through $d_j^r$
encodes exclusion. The sharp set follows from two linear programs
\citep{balke1997bounds}:
\begin{equation}
\begin{aligned}
    \Theta_I^{(\mathrm{IVb})}(x)
      &=[L_{\mathrm{IV}}(x),U_{\mathrm{IV}}(x)],\\
    L_{\mathrm{IV}}(x)
      &=\min_{\pi\in\Pi_x}\sum_{r\in\mathcal R}(y_1^r-y_0^r)\pi_r,\\
    U_{\mathrm{IV}}(x)
      &=\max_{\pi\in\Pi_x}\sum_{r\in\mathcal R}(y_1^r-y_0^r)\pi_r.
\end{aligned}
\label{eq:appB_iv_bounds_set}
\end{equation}
No outcome monotonicity or generator-specific gain restriction is
added to these programs. For compatible population distributions,
$\Theta_I^{(\mathrm{IVb})}(x)\subseteq\Theta_I^{(\mathrm M)}(x)$
on the matched companion views.

\subsubsection{Sensitivity bounds}
Let $e(x,u):=P(T=1\mid X=x,U=u)$ and
$e(x):=P(T=1\mid X=x)$. The marginal sensitivity model (MSM) maintains, for a
specified $1\leq\Gamma<\infty$,
\begin{equation}
    \Gamma^{-1}
    \leq
    \frac{e(x,u)/\{1-e(x,u)\}}{e(x)/\{1-e(x)\}}
    \leq\Gamma.
    \label{eq:appB_msm_restriction}
\end{equation}
The compatible causal laws satisfy consistency, positivity, and
latent exchangeability
$\{Y(0),Y(1)\}\perp T\mid X,U$ and reproduce the observed
$(X,T,Y)$ distribution.
The latent variable in this nonparametric compatible-law class is
otherwise unrestricted; the generator's binary latent support and
within-$X$ gain invariance are not imposed \citep{oprescu2023b}.

\paragraph{The MSM identified set.}
We specialize the CATE identified set in
Appendix~\ref{app:identification_terminology} to these assumptions.
For $t\in\{0,1\}$, write $p_t(x):=P(T=t\mid X=x)$ and let
$F_t(\cdot\mid x)$ denote the observed law of $Y\mid T=t,X=x$,
with a finite absolute first moment.
The odds restriction induces the normalized reweighting class
\begin{equation*}
\begin{aligned}
    \ell_t^\Gamma(x)&:=p_t(x)+\{1-p_t(x)\}/\Gamma,\\
    u_t^\Gamma(x)&:=p_t(x)+\Gamma\{1-p_t(x)\},\\
    \mathcal W_t^\Gamma(x)
      &:=\left\{w:\ \ell_t^\Gamma(x)\leq w(y)\leq u_t^\Gamma(x)
        \quad F_t\text{-a.s.},\quad
        \int w(y)\,dF_t(y\mid x)=1\right\}.
\end{aligned}
\end{equation*}
The extremal potential-outcome means are
\begin{equation*}
\begin{aligned}
    \underline\mu_t^\Gamma(x)
      &:=\inf_{w\in\mathcal W_t^\Gamma(x)}
          \int y\,w(y)\,dF_t(y\mid x),\\
    \overline\mu_t^\Gamma(x)
      &:=\sup_{w\in\mathcal W_t^\Gamma(x)}
          \int y\,w(y)\,dF_t(y\mid x).
\end{aligned}
\end{equation*}
The sharp CATE set under this model is
\citep{oprescu2023b}
\begin{equation}
\begin{aligned}
    \Theta_I^{(\mathrm{HC})}(x;\Gamma)
      &=[L_\Gamma(x),U_\Gamma(x)],\\
    L_\Gamma(x)
      &=\underline\mu_1^\Gamma(x)-\overline\mu_0^\Gamma(x),\\
    U_\Gamma(x)
      &=\overline\mu_1^\Gamma(x)-\underline\mu_0^\Gamma(x).
\end{aligned}
    \label{eq:appB_msm_set}
\end{equation}
Normalization makes each reweighted outcome law a probability law.
At $\Gamma=1$, all weights equal one and the set reduces to the
observed treatment-group mean contrast.
Larger $\Gamma$ allows greater hidden selection and yields nested,
nonshrinking population sets.
The set contains $\tau_0(x)$ when the true law satisfies the maintained
assumptions at the specified $\Gamma$.

\section{SCM and Causal Worlds}
\label{app:master_scm}

This appendix describes the synthetic SCM used in Table~\ref{tab:observational_views}, the parameter and noise sampling procedure and the
construction of paired observational views. The main comparison uses
40 worlds, indexed by $w\in\{0,\ldots,39\}$, each containing 1,024
context units and 100 query units.

\subsection{Full SCM}
\label{app:full_master_scm}

\paragraph{Causal world and structure.}
Each world describes the effect of a binary treatment $T\in\{0,1\}$
on a continuous outcome $Y$, mediated entirely through a continuous
post-treatment variable $M$. Pre-treatment covariates $X\in\mathbb R^{50}$
and a confounder $U\in\{-1,+1\}$ influence both treatment assignment
and the outcome, with $U$ inducing the back-door path
$T\leftarrow U\to Y$. A binary instrument $I\in\{0,1\}$ also
affects treatment assignment but influences the outcome only through
the directed path $I\to T\to M\to Y$. $X$, $U$, and $I$
form the roots of the SCM. The pre-treatment proxies
$(Z_p,W_p)\in\{0,1\}^2$ are generated from $(X,U)$, providing noisy
measurements of the confounder without directly affecting treatment
or outcome.

The full world retains these variables together,
\[
    \mathcal W
    =
    \{X,U,I,Z_p,W_p,T,M(0),M(1),M,Y(0),Y(1),Y\}.
\]

\subsection{SCM Generation Algorithm}
\label{app:master_scm_algorithm}

Algorithm~\ref{alg:master_scm_generation} summarizes how a single
causal world is generated and converted into paired observational
views. World-level mechanisms are sampled and calibrated first,
followed by unit-level realizations and a shared context--query split.
The structural equations and parameter settings are detailed in
Sections~\ref{app:structural_equations}
and~\ref{app:mechanism_distributions}.

\begin{algorithm}[t]
\caption{SCM generation and multi-view construction}
\label{alg:master_scm_generation}
\small
\algrenewcommand{\algorithmicrequire}{\textbf{Input:}}
\algrenewcommand{\algorithmicensure}{\textbf{Output:}}
\begin{algorithmic}[1]

\Require World seed $s_w$; settings in Table~\ref{tab:appC_settings}
\Ensure Full world $\mathcal W_w$, paired views
$\{\mathcal O_w^{(r)}\}_r$, and shared split
$(\mathcal C_w,\mathcal Q_w)$

\State Select the treatment-effect family using $s_w$ and sample
its parameters and the remaining SCM parameters.

\State Calibrate $\tau_w(x)$ and the treatment intercept $b_{T,w}$
on an auxiliary covariate sample; set
$\beta(x)\gets\tau_w(x)/\delta$.

\For{$i=1,\ldots,N_c+N_q$}

    \State Draw root variables $(X_i,U_i,I_i)$ and independent
    exogenous noises for treatment, proxies, mediator, and outcome.

    \State Generate $T_i^I(0),T_i^I(1)$ using the same treatment
    threshold; set $T_i\gets T_i^I(I_i)$.

    \State Generate $(Z_{p,i},W_{p,i})$ conditionally independently
    given $(X_i,U_i)$.

    \State Construct paired potentials using the same
    $\epsilon_{M,i}$ and $\epsilon_{Y,i}$ in both arms:
    \Statex \hspace{\algorithmicindent}
    $\begin{aligned}
        M_i(t)&\gets\delta t+\sigma_M\epsilon_{M,i},\\
        Y_i(t)&\gets\mu(X_i)+\gamma_U(X_i)U_i
                   +\beta(X_i)M_i(t)+\sigma_Y\epsilon_{Y,i},
        \qquad t\in\{0,1\}.
    \end{aligned}$

    \State Select the factual values:
    $M_i\gets M_i(T_i)$ and $Y_i\gets Y_i(T_i)$.

\EndFor

\State Assemble $\mathcal W_w$ from the generated variables
and potential outcomes.

\State Randomly partition the unit indices into disjoint sets
$\mathcal C_w$ and $\mathcal Q_w$, with
$|\mathcal C_w|=N_c$ and $|\mathcal Q_w|=N_q$.

\For{$r\in\{\mathrm{BD},\mathrm{FD},\mathrm{IV},\mathrm{PX},\mathrm{HC}\}$}

    \State Project the same full world onto the variables
    retained in view $r$:
    \Statex \hspace{\algorithmicindent}
    $\displaystyle
        \mathcal O_w^{(r)}
        \gets
        \left\{(X_i,T_i,Y_i,V_i^{(r)})\right\}_{i=1}^{N_c+N_q}.
    $

    \State Reuse $(\mathcal C_w,\mathcal Q_w)$ and the query targets
    $\{\tau_w(X_i):i\in\mathcal Q_w\}$ without resampling.

\EndFor

\end{algorithmic}
\end{algorithm}

\subsection{Structural Equations and Design Choices}
\label{app:structural_equations}

We suppress the world index when describing the structural equations
for a single realized world.

\paragraph{Root variables.}
The root distributions are
\begin{equation*}
    X_i\sim\mathcal N(0,I_{50}),\qquad
    P(U_i=-1)=P(U_i=+1)=\tfrac12,\qquad
    I_i\sim\operatorname{Bernoulli}(\tfrac12).
\end{equation*}

\paragraph{Treatment assignment.}
For $j\in\{0,1\}$, define
\begin{align*}
    e_j(x,u)
       &=\operatorname{expit}\{b_T+0.55x^\top v_T+0.75u+1.50j\},\\
    T_i^I(j)
       &=\mathbf 1\{V_{T,i}\le e_j(X_i,U_i)\},
          \qquad V_{T,i}\sim\operatorname{Uniform}(0,1),\\
    T_i
       &=T_i^I(I_i).
\end{align*}
Because the same threshold $V_{T,i}$ is used for both instrument
values and the coefficient on $j$ is positive, the construction
satisfies $T_i^I(1)\ge T_i^I(0)$.

\paragraph{Proxies.}
The treatment-inducing and outcome-inducing proxies are generated by
\begin{align*}
    Z_{p,i}\mid X_i,U_i
      &\sim \operatorname{Bernoulli}
      \!\left[\operatorname{expit}\{0.20X_i^\top v_Z+2U_i\}\right],\\
    W_{p,i}\mid X_i,U_i
      &\sim \operatorname{Bernoulli}
      \!\left[\operatorname{expit}\{0.20X_i^\top v_W+2U_i\}\right].
\end{align*}
They are noisy pre-treatment measurements of the latent confounder,
not direct causes of treatment or outcome.

\paragraph{Mediator and outcomes.}
For $t\in\{0,1\}$,
\begin{equation*}
    M_i(t)=t+0.75\epsilon_{M,i},
    \qquad
    M_i=M_i(T_i),
    \qquad
    \epsilon_{M,i}\sim\mathcal N(0,1).
\end{equation*}
The prognostic surface and outcome-confounding coefficient are
\begin{align*}
    \mu(x)
      &=0.7x^\top v_\mu
        +0.25\sin\{x^\top \operatorname{roll}(v_\mu,1)\},\\
    \gamma_U(x)
      &=0.8\{1+0.75\tanh(x^\top v_\gamma)\}.
\end{align*}
Let $\tau_0(x)$ denote the calibrated treatment-effect surface and set $
\beta(x)=\tau_0(x)$.

The potential outcomes are
\begin{equation*}
    Y_i(t,m)
    =
    \mu(X_i)+\gamma_U(X_i)U_i+\beta(X_i)m+0.75\epsilon_{Y,i},
    \qquad
    \epsilon_{Y,i}\sim\mathcal N(0,1),
\end{equation*}
and
\begin{equation*}
    Y_i(t)=Y_i\{t,M_i(t)\},
    \qquad
    Y_i=T_iY_i(1)+(1-T_i)Y_i(0).
\end{equation*}

\paragraph{Why the target is shared across views.}
Because the same mediator noise and the same outcome noise are shared
across the two treatment arms, we obtain
\begin{equation*}
    Y_i(1)-Y_i(0)
    =
    \beta(X_i)\{M_i(1)-M_i(0)\}
    =
    \tau_0(X_i).
\end{equation*}
Hence the benchmark target is identical across sibling observational
views of the same realized world. Changing the observational view
changes only which auxiliary variables are observed and which
identification assumptions are available to the estimator; it does not
change the underlying query units or the target effect values.

\subsection{Parameterization and Fixed Settings}
\label{app:mechanism_distributions}

\paragraph{Sparse directions.}
Each coefficient direction is sampled as a sparse unit vector in
$\mathbb R^{50}$. The number of active coordinates is fixed by role:
the CATE directions use 10 active coordinates each,
$v_\mu$ and $v_T$ use 14,
$v_Z$ and $v_W$ use 8,
and $v_\gamma$ uses 10.

\paragraph{Treatment-effect families.}
Let $a=x^\top v_a$ and $b=x^\top v_b$. The uncalibrated effect surface
$q_w(x)$ is chosen from one of the following five families:
linear, quadratic-interaction, threshold/piecewise,
Fourier/nonstationary, and shallow MLP.
The world seed determines the family so that the 40 worlds contain
eight worlds from each family.

\paragraph{Calibration.}
An auxiliary Gaussian sample of size $N_{\mathrm{cal}}=50{,}000$ is
used to calibrate
\[
\tau_w(x)=0.5+0.5\frac{q_w(x)-\bar q_w}{s_{q,w}},
\]
where $\bar q_w$ and $s_{q,w}$ are the empirical mean and standard
deviation of $q_w$ on the calibration sample. The same auxiliary
sample is also used to calibrate the treatment intercept $b_T$ by
bisection so that the average treatment propensity is approximately
$0.5$.

\begin{table}[htbp]
\centering
\caption{Main SCM numerical configuration.}
\label{tab:appC_settings}
\small
\setlength{\tabcolsep}{4pt}
\renewcommand{\arraystretch}{1.08}

\begin{tabular}{@{}p{0.67\linewidth}l@{}}
\toprule
\textbf{Quantity} & \textbf{Value} \\
\midrule

Number of world-generation seeds
& 40 \\

Baseline dimension
& 50 \\

Context / query units per world
& 1,024 / 100 \\

Calibration sample size
& 50,000 \\

Calibration CATE mean / standard deviation
& 0.5 / 0.5 \\

Treatment covariate / latent / instrument loading
& 0.55 / 0.75 / 1.50 \\

Outcome-confounding base / modulation
& 0.80 / 0.75 \\

Mediator shift / mediator noise / outcome noise
& 1.0 / 0.75 / 0.75 \\

Proxy covariate / latent loading
& 0.20 / 2.00 \\

Severity multiplier $\kappa$
& 1.0 \\

Stored oracle samples per arm/query
& 8,192 \\

\bottomrule
\end{tabular}
\end{table}

\subsection{Full SCM for Partial Identification}
\label{app:partial_id_scm}

Compared with the point-estimation SCM, the partial-identification
companion omits the mediator and proxies and replaces the continuous
outcome mechanism with binary potential outcomes.
It retains Gaussian covariates, a binary latent confounder and
instrument, and shared-threshold logistic treatment assignment,
but generates 40 new worlds with separately calibrated parameters.

Algorithm~\ref{alg:partial_id_scm} summarizes the construction.
The outcome amplitudes satisfy
$A_b,A_\tau>0$ and $A_b+A_\tau/2\leq0.45$, ensuring
$p_t(x,u)\in[0.05,0.95]$ without clipping.
Since $p_1(x,u)-p_0(x,u)=\tau_0(x)$, the target remains
$\mathbb E[Y(1)-Y(0)\mid X=x]=\tau_0(x)$, but the realized
unit-level effect need not equal this conditional mean.
Hidden-confounding strength is calibrated on separate development
seeds, with the MSM restriction at $\Gamma=2$ required to hold
after marginalizing over the instrument.
These generator-specific restrictions are not added to the
identified-set assumptions in Appendix~\ref{app:hidden_confounding}.

\begin{algorithm}[t]
\caption{Partial-identification SCM and matched-view construction}
\label{alg:partial_id_scm}
\small
\begin{algorithmic}[1]
\Require World seed; calibrated parameters with $a_I>0$;
         $N_c=1024$, $N_q=100$
\Ensure Matched contexts, shared query covariates and CATE targets,
        and population-oracle bounds

\State Select one of five effect families and sample its raw
       function $q_f$, baseline function $f_b$, and unit-normalized
       treatment direction $v_T$.
\State $e_j(x,u)\gets
       \operatorname{expit}(b_T+a_Xx^\top v_T+a_Uu+a_Ij)$,
       $j\in\{0,1\}$
\State $b(x,u)\gets \frac12+A_b\tanh\{f_b(x)+c_Uu\}$
\State $\tau_0(x)\gets A_\tau\tanh\{q_f(x)\}$

\For{$i=1,\ldots,N_c+N_q$}
    \State Draw independent $X_i\sim\mathcal N(0,I_{50})$,
           $U_i\sim\operatorname{Unif}\{-1,+1\}$,
           and $I_i\sim\operatorname{Bernoulli}(1/2)$.
    \State Draw independent
           $V_{T,i},V_{Y,i}\sim\operatorname{Uniform}(0,1)$.
    \State $T_i^I(j)\gets
           \mathbf 1\{V_{T,i}\leq e_j(X_i,U_i)\}$,
           $j\in\{0,1\}$; \quad $T_i\gets T_i^I(I_i)$
    \State $p_{it}\gets b(X_i,U_i)+(t-\frac12)\tau_0(X_i)$,
           $t\in\{0,1\}$
    \State $Y_i(t)\gets\mathbf 1\{V_{Y,i}\leq p_{it}\}$,
           $t\in\{0,1\}$; \quad $Y_i\gets Y_i(T_i)$
\EndFor

\State Assign one disjoint context--query split
       $(\mathcal C,\mathcal Q)$.
\State $\mathcal D_{\mathrm M}
       =\mathcal D_{\mathrm S}
       \gets\{(X_i,T_i,Y_i):i\in\mathcal C\}$
\State $\mathcal D_{\mathrm{IV}}
       \gets\{(X_i,I_i,T_i,Y_i):i\in\mathcal C\}$
\State Reuse $\{X_i:i\in\mathcal Q\}$ and
       $\{\tau_0(X_i):i\in\mathcal Q\}$ across all three settings.
\State Compute population probabilities analytically at query
       covariates and apply the Manski, monotone-IV, and binary-MSM
       bound functionals in Appendix~\ref{app:hidden_confounding}.
\end{algorithmic}
\end{algorithm}

Manski and sensitivity analyses use identical observational data
but different maintained assumptions. IV additionally observes $I$.
All three settings preserve the realized $(X,T,Y)$, split, query
units, and target CATE.
Sensitivity oracles use
$\Gamma\in\{1,1.25,1.5,2,3,5\}$.
Oracle quantities are reserved for evaluation and are not supplied
to estimators.

\section{Data Validation and Construction Audits}
\label{sec:data-validation}

We audit the 40 saved synthetic worlds, each containing 1,024
context and 100 query units.
All checks use existing benchmark artifacts and evaluator-only
SCM quantities, without regenerating data or running estimators.
Production-input hashes match before and after the audit.
Copied quantities are compared by exact equality, whereas
deterministic numerical identities are checked using an absolute
tolerance of $10^{-10}$.

\subsection{Construction Integrity}
\label{app:construction-integrity}

\paragraph{Paired-view integrity.}
We compare the five observational views within each world,
covering 200 view tables and 400 within-world view pairs.
The comparisons include world and unit identifiers, $X$, $T$, $Y$,
context--query membership, ordered query identifiers, and stored
query targets. Each view is also checked against its saved master-world
projection, including its retained auxiliary variables. Pairing is verified from stored values and row identities,
rather than inferred from shared generation seeds.

\paragraph{Target consistency.}
Using the saved potential outcomes and targets, we evaluate
$Y_i(1)-Y_i(0)-\tau_0(X_i)$ over all 44,960 units.
The maximum absolute discrepancy is
$1.55\times10^{-15}$, with no violations of the
$10^{-10}$ tolerance. This verifies the benchmark-specific identity that each unit’s treatment effect equals the target CATE evaluated at its covariates.

\paragraph{Stay/Move targets.}
We check the evaluation targets stored in the completed Stay and Move experiment.
Repeated target entries across estimator--regime configurations
agree exactly, and all conditions share the same ordered
query identifiers.
The Original targets exactly match the production-world targets.
Across the 4,000 paired queries, the saved targets satisfy
$\tau_0^{\mathrm{S}}(X_i)=\tau_0(X_i)$ and
$\tau_0^{\mathrm{M}}(X_i)=2\tau_0(X_i)$
with zero discrepancy.
Table~\ref{tab:audit-construction-integrity} summarizes these results.

\begin{table}[t]
\centering
\caption{
\textbf{Construction integrity.}
Pairing checks cover all 40 worlds and their five observational
views.
The potential-outcome target check uses all 44,960 units. Saved Stay/Move target checks use 4,000 paired queries. Maximum errors are taken over all applicable worlds and units.
}
\label{tab:audit-construction-integrity}
\small
\setlength{\tabcolsep}{5pt}
\renewcommand{\arraystretch}{1.12}
\begin{tabular*}{\linewidth}{
    @{\extracolsep{\fill}}lr@{}
}
\toprule
Check & Result \\
\midrule
Shared-field mismatches
    & $0$ \\
Master-projection mismatches
    & $0$ \\
Query-order mismatches
    & $0$ \\
Target identity:
$\max_i |Y_i(1)-Y_i(0)-\tau_0(X_i)|$
    & $1.55\times10^{-15}$ \\
Saved Stay target:
$\max_i |\tau_0^{\mathrm{S}}(X_i)
          -\tau_0(X_i)|$
    & $0$ \\
Saved Move target:
$\max_i |\tau_0^{\mathrm{M}}(X_i)
          -2\tau_0(X_i)|$
    & $0$ \\
\bottomrule
\end{tabular*}
\end{table}

\subsection{Regime-wise Functional Consistency and Non-degeneracy}
\label{app:regime-validation}

\paragraph{Analytic functional consistency.}
At the same 4,000 query locations, we reconstruct the CATE
through back-door adjustment, the front-door functional,
the conditional Wald ratio, and the proximal bridge functional.
These calculations use saved structural quantities and
SCM-derived conditional means and probabilities.
The stored targets are used only for comparison after
reconstruction, not as inputs to the identifying functionals.
All reconstructed values are finite, and no absolute discrepancy
exceeds $10^{-10}$.
Table~\ref{tab:audit-regime-validation} reports the RMSE and
maximum absolute error for each regime.
These results establish numerical consistency between the
analytic SCM-based functionals and the stored targets.

\paragraph{Numerical construction diagnostics.}
We reproduce the existing certification gates using their
original evaluation populations and aggregation rules.
Complete treatment propensities
$P(T=1\mid X,U,I)$ are evaluated at every stored $(X,U)$ pair
under both instrument values, giving 2,248 probabilities
per world.
These are distinguished from the marginal back-door propensity
$P(T=1\mid X,U)$.
The front-door diagnostics are computed from the structural
mediator shift $\delta$ and noise scale $\sigma_M$.

For IV, we compute
$
\Delta_T(x)
=
E[T\mid I=1,X=x]-E[T\mid I=0,X=x]
$
at all generated covariate locations per world and
apply the gate to the within-world fifth percentile.
We evaluate the bridge operator 
\[
A_t(x)_{z,w}
=
P(W_p=w\mid Z_p=z,T=t,X=x),
\qquad z,w\in\{0,1\},
\]
at the same covariate locations under both treatment arms in PX.
We record each world's minimum singular value and maximum
condition number.
All 40 worlds satisfy the recorded gates on these populations.

\paragraph{Hidden-confounding strength.}
For the confounding-bias function $b_{\mathrm{HC}}(x)$
defined in Appendix~B, we compute
\[
R_{\mathrm{HC},w}
=
\frac{
    \operatorname{SD}_{Q_w}\!\left[b_{\mathrm{HC}}(X)\right]
}{
    \operatorname{SD}_{Q_w}\!\left[\tau_0(X)\right]
},
\]
where both standard deviations are computed over the 100 queries
in world $w$ using divisor $N_q$.
The ratio ranges from $0.3699$ to $0.5999$, and every world
passes the recorded threshold of $0.35$.
This measures covariate-varying distortion of the observational
contrast relative to CATE variation.
HC is not assigned an oracle point-identification reconstruction.

\begin{table}[t]
\centering
\caption{
\textbf{Regime diagnostics and analytic functional consistency.}
Diagnostics use the certification populations described in the text.
Bracketed values give ranges of world-level statistics, except
for complete propensity, which gives the global range over
checked probabilities.
FD diagnostics are identical across worlds.
Oracle reconstruction uses 4,000 shared queries per regime.
}
\label{tab:audit-regime-validation}
\small
\setlength{\tabcolsep}{4pt}
\renewcommand{\arraystretch}{1.12}

\begin{tabular*}{\linewidth}{
    @{\extracolsep{\fill}}p{0.46\linewidth}cr@{}
}
\toprule
Diagnostic & Criterion & Result \\
\midrule
Complete propensity $P(T=1\mid X,U,I)$
    & $[0.01,0.99]$
    & $[0.02998,0.98115]$ \\

FD standardized shift $|\delta|/\sigma_M$
    & $\geq 0.75$
    & $1.3333$ \\

FD support score $\exp\{-\delta^2/(8\sigma_M^2)\}$
    & $\geq 0.60$
    & $0.8007$ \\

IV first-stage fifth percentile
    & $\geq 0.10$
    & $[0.2601,0.2737]$ \\

PX minimum singular value
    & $\geq 0.01$
    & $[0.4139,0.4655]$ \\

PX maximum condition number
    & $\leq 250$
    & $[2.2484,2.5821]$ \\

HC bias-to-effect SD ratio $R_{\mathrm{HC}}$
    & $\geq 0.35$
    & $[0.3699,0.5999]$ \\
\bottomrule
\end{tabular*}

\vspace{6pt}

\begin{tabular*}{\linewidth}{
    @{\extracolsep{\fill}}lcrr@{}
}
\toprule
Oracle functional & Queries & RMSE & Max.\ abs.\ error \\
\midrule
Back-door
    & 4,000
    & $7.75\times10^{-17}$
    & $4.44\times10^{-16}$ \\

Front-door
    & 4,000
    & $1.05\times10^{-16}$
    & $6.66\times10^{-16}$ \\

Conditional Wald
    & 4,000
    & $2.85\times10^{-16}$
    & $1.67\times10^{-15}$ \\

Proximal bridge
    & 4,000
    & $1.91\times10^{-16}$
    & $8.88\times10^{-16}$ \\
\bottomrule
\end{tabular*}
\end{table}

\section{Experimental Setup}
\label{sec:experimental_setup}
\subsection{Metrics}
\subsubsection{Point-Effect and Heterogeneity Metrics}
\label{app:evaluation_metrics}

We specify the metrics and aggregation rules used in Figures~\ref{fig:regime_point_estimation} and~\ref{fig:structural_response}.
Metrics are first computed within each held-out query set and then
summarized across paired worlds or IHDP realizations. The synthetic
comparisons use 40 worlds with 100 queries per world; the IHDP comparison
uses 100 realizations with 75 test queries each. 

Our point-error metrics from Section ~\ref{sec:experiments} are
\begin{align*}
    \mathrm{sPEHE}
       &=\sqrt{
           \frac{1}{N_q}\sum_{i=1}^{N_q}
           \bigl(\widehat{\tau}(x_i)-\tau(x_i)\bigr)^2
         },
    \\
    \mathrm{ATE\ error}
       &=\left|
           \widehat{\tau}_{\mathrm{ATE},Q}
           -\tau_{\mathrm{ATE},Q}
         \right|,
    \\
    \mathrm{c\text{-}sPEHE}
       &=\left[
           \frac{1}{N_q}\sum_{i=1}^{N_q}
           \left\{
             \bigl(\widehat{\tau}(x_i)
                    -\widehat{\tau}_{\mathrm{ATE},Q}\bigr)
             -
             \bigl(\tau(x_i)-\tau_{\mathrm{ATE},Q}\bigr)
           \right\}^2
         \right]^{1/2}.
\end{align*}

$\tau(x)$ denotes the oracle CATE target for the evaluation
under consideration. $\tau_0(x)$ for the main synthetic benchmark
and $\tau_{\mathrm{true}}(X,U)$ for IHDP.

\paragraph{Exact decomposition.}
Write $e_i=\widehat{\tau}(x_i)-\tau(x_i)$ and define its query mean as
\begin{equation*}
    \bar e
       =\frac{1}{N_q}\sum_{i=1}^{N_q}e_i
       =\widehat{\tau}_{\mathrm{ATE},Q}
          -\tau_{\mathrm{ATE},Q}.
\end{equation*}
Since the centered residuals sum to zero,
\begin{align*}
    \frac{1}{N_q}\sum_{i=1}^{N_q}e_i^2
       &=\frac{1}{N_q}\sum_{i=1}^{N_q}
           \bigl\{(e_i-\bar e)+\bar e\bigr\}^2
         \\
       &=\frac{1}{N_q}\sum_{i=1}^{N_q}(e_i-\bar e)^2
         +2\bar e\underbrace{
             \frac{1}{N_q}\sum_{i=1}^{N_q}(e_i-\bar e)
           }_{0}
         +\bar e^2.
\end{align*}
Consequently, the decomposition follows:
\begin{equation*}
    \boxed{
        \mathrm{sPEHE}^2
        =
        \bigl(
            \widehat{\tau}_{\mathrm{ATE},Q}
            -
            \tau_{\mathrm{ATE},Q}
        \bigr)^2
        +
        \mathrm{c\text{-}sPEHE}^2
    }.
\end{equation*}
All terms use the same queries and uniform weights. The first term
measures the global mean-effect discrepancy, whereas the second
measures error in the centered CATE function. A smaller total error
alone therefore does not determine which component improved.
Figure~\ref{fig:regime_point_estimation} stacks the world means of these two squared terms, rather
than their square roots or centered quantities computed after pooling
queries across worlds.

\subsubsection{Sensitivity Metrics}
Let $\widetilde{\mathcal I}_{wi}^{\Gamma}
=[\widetilde L_{wi}^{\Gamma},\widetilde U_{wi}^{\Gamma}]$ denote the stored
oracle-reference sensitivity interval. Let $n_w := |Q_w|$ denote the number of query units in world $w$.
For predicted endpoints $(\widehat L_{wi}^{\Gamma},
\widehat U_{wi}^{\Gamma})$, the combined endpoint error is
\begin{equation*}
    \operatorname{EndpointRMSE}_w(\Gamma)
      =\left[\frac{1}{2n_w}\sum_{i\in\mathcal Q_w}
        \left\{(\widehat L_{wi}^{\Gamma}-\widetilde L_{wi}^{\Gamma})^2
             +(\widehat U_{wi}^{\Gamma}-\widetilde U_{wi}^{\Gamma})^2
        \right\}\right]^{1/2}.
\end{equation*}

We additionally report the fraction of query-unit oracle effects contained
within the estimated bounds:
\begin{equation*}
    \operatorname{Containment}_w(\Gamma)
      = \frac{1}{n_w}\sum_{i\in\mathcal Q_w}
        \mathbf{1}\!\left\{
            \widehat L_{wi}^{\Gamma}
            \leq \tau_{wi}
            \leq \widehat U_{wi}^{\Gamma}
        \right\}.
\end{equation*}

\subsection{Baselines}
\label{sec:baselines}
\subsubsection{Causal foundation models}
\begin{figure}[t]
    \centering
    \includegraphics[width=\linewidth]{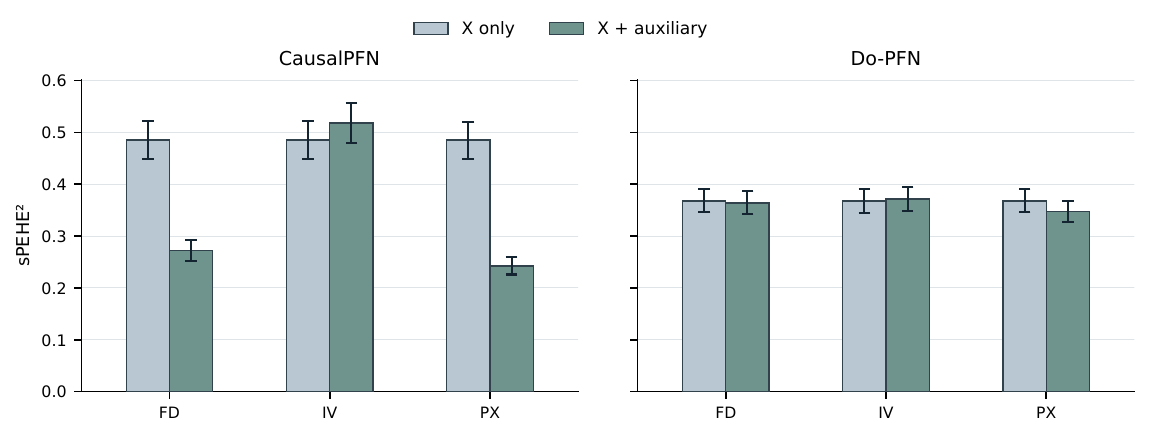}
    \caption{
    \textbf{Effect of auxiliary-variable inputs for CausalPFN and Do-PFN.}
    We compare $X$-only inputs with inputs augmented by the
    regime-specific auxiliary variables in the front-door (FD),
    instrumental-variable (IV), and proximal (PX) views. The auxiliary-input FD configurations condition on the factual mediator and are included as input-sensitivity diagnostics, not as front-door estimators of the total effect.
    }
    \label{fig:cfm-input-ablation}
\end{figure}

We evaluate CausalPFN, Do-PFN, and CausalFM
\citep{balazadeh2026causalpfn,robertson2026pfn,ma2026foundation}
using their released checkpoints without fine-tuning.
CausalPFN and CausalFM return CATE predictions directly,
whereas Do-PFN contrasts predictions under the two treatment
values. 
FD configurations of CausalPFN and Do-PFN omit $M$ from both context and query to avoid conditioning total-effect predictions on a factual
mediator. They are mediator-free transfer evaluations,
not front-door estimators. Figure~\ref{fig:cfm-input-ablation} compares these models under
$X$-only inputs and inputs augmented with the regime-specific
auxiliary variables, illustrating the sensitivity of their
predictions to the input configuration.
CausalFM uses its native FD and IV checkpoints in the
corresponding regimes while CausalPFN and Do-PFN receive the auxiliary
variables as ordinary covariates without role annotations
or identification wrappers.

These comparisons concern the evaluated model–input configurations, not equally supported native estimators in every regime.

\subsubsection{Modular approaches}

Our modular estimators combine supervised predictive models with
explicit causal estimation procedures matched to each identification
regime.
The causal procedure determines which nuisance quantities are required
and how their predictions are combined into a CATE estimate, while the
predictive backbone estimates the corresponding conditional means,
probabilities, or pseudo-outcome regressions.
When multiple predictive backbones are evaluated within the same
procedure, we keep the causal estimation step fixed and change only
the supervised models used for these predictive components.

Depending on the regime, we instantiate the predictive components with
TabPFN-v3.5 \citep{jager2026tabpfn}, XGBoost
\citep{chen2016xgboost}, feed-forward neural networks, or ExtraTrees
\citep{geurts2006extremely}.
For TabPFN-v3.5, continuous targets use
\texttt{TabPFNRegressor} and binary probabilities use
\texttt{TabPFNClassifier}. We implement XGBoost with 300 estimators, maximum tree depth 3,
and learning rate 0.05, using the histogram tree method.
The neural-network backbone is a two-layer MLP with 128 hidden units
per layer and ReLU activations.
It is trained with Adam using a learning rate of $10^{-3}$,
weight decay $10^{-4}$, and batch size 128.
ExtraTrees uses 300 trees with
\texttt{max\_features}=0.8. The minimum leaf size is 8 for classification, 6 for nuisance
regression, and 10 for the final CATE regression.
All remaining parameters use the corresponding package defaults
and we perform no hyperparameter optimization.

\paragraph{Back-door (BD).}
We evaluate S-, X-, and DR-Learners using the observed adjustment
vector $V=(X,U)$.
The S-Learner \citep{kunzel2019metalearners} fits a single outcome regression
$\widehat g(V,T)$ on the full context and returns
$\widehat g(v,1)-\widehat g(v,0)$.

The X-Learner \citep{kunzel2019metalearners} uses five-fold
treatment-stratified cross-fitting to obtain arm-specific
out-of-fold outcome predictions.
For controls, it constructs
$D_i^0=\widehat m_{1,-k(i)}(V_i)-Y_i$,
whereas for treated observations it uses
$D_i^1=Y_i-\widehat m_{0,-k(i)}(V_i)$.
Separate regressions of $D^0$ and $D^1$ produce
$\widehat f_0(v)$ and $\widehat f_1(v)$.
A propensity model fitted on the full context estimates
$\widehat e(v)=\widehat P(T=1\mid V=v)$.
The final X-Learner estimate is
\[
\widehat\tau_{\mathrm{X}}(v)
=
\widehat e(v)\widehat f_0(v)
+
\{1-\widehat e(v)\}\widehat f_1(v).
\]

The DR-Learner \citep{kennedy2023towards} cross-fits both
arm-specific outcome models and the propensity model.
It constructs the doubly robust pseudo-outcome
\[
\Gamma_i
=
\widehat m_1(V_i)-\widehat m_0(V_i)
+
\frac{T_i\{Y_i-\widehat m_1(V_i)\}}{\widehat e(V_i)}
-
\frac{(1-T_i)\{Y_i-\widehat m_0(V_i)\}}
     {1-\widehat e(V_i)},
\]
and fits a final regression of $\Gamma_i$ on $V_i$.

All three estimators receive the factual query $U$ and do not
marginalize over it.
Their evaluation against $\tau_0(X)$ relies on the benchmark
restriction
$\mathbb{E}[Y(1)-Y(0)\mid X,U]=\tau_0(X)$.

\paragraph{Front-door (FD).}
We use a common front-door plug-in procedure~\citep{pearl2022causal, guo2023flexible, ma2026foundation} with neural-network,
XGBoost, or TabPFN-v3.5 predictive components.
Across these variants, the front-door functional and numerical
integration procedure are identical. Only the supervised models used
for the treatment, mediator, and outcome regressions change.
Algorithm~\ref{alg:fd-plugin} summarizes the estimator.

\begin{algorithm}[t]
\caption{Front-door plug-in estimator}
\label{alg:fd-plugin}
\begin{algorithmic}[1]
\Require Context data $\{(X_i,T_i,M_i,Y_i)\}_{i=1}^{n}$, query covariate $x$
\Ensure Front-door estimate $\widehat\tau_{\mathrm{FD}}(x)$
\State Fit treatment model $\widehat e(x)\approx P(T=1\mid X=x)$
\State Set $\widehat p_1(x):=\widehat e(x)$ and
$\widehat p_0(x):=1-\widehat e(x)$
\State Fit outcome model $\widehat Q(x,m,s)\approx \mathbb{E}[Y\mid X=x,M=m,T=s]$
\For{$a\in\{0,1\}$}
    \State Split the context observations with $T_i=a$ into five folds
    \For{each fold $k$}
        \State Fit $\widehat g_{a,-k}(x)\approx \mathbb{E}[M\mid T=a,X=x]$ on the other four folds
        \State Predict the held-out fold and compute raw residuals
        \[
        r^{\mathrm{raw}}_{ai}
        =
        M_i-\widehat g_{a,-k(i)}(X_i),
        \qquad T_i=a
        \]
    \EndFor
    \State Center residuals within arm $a$:
    \[
    r_{ai}
    =
    r^{\mathrm{raw}}_{ai}
    -
    \frac{1}{n_a}\sum_{j:T_j=a} r^{\mathrm{raw}}_{aj}
    \]
    \State Refit $\widehat g_a(x)\approx \mathbb{E}[M\mid T=a,X=x]$ on all observations with $T_i=a$
    \State Form the arm-specific mediator support $M_a(x)=\{\widehat g_a(x)+r_{ai}:T_i=a\}$
    \State Compute
    $
    \widehat\theta_a(x)
    =
    \frac{1}{|\mathcal M_a(x)|}
    \sum_{m\in\mathcal M_a(x)}
    \sum_{s=0}^{1}
    \widehat p_s(x)\widehat Q(x,m,s)
    $
\EndFor
\State \Return
$
\widehat\tau_{\mathrm{FD}}(x)
=
\widehat\theta_1(x)-\widehat\theta_0(x)
$
\end{algorithmic}
\end{algorithm}

Our implementation approximates the conditional mediator distribution
in each treatment arm by shifting the complete pool of centered
out-of-fold mediator residuals around the predicted mediator mean.
Thus, rather than fitting a separate conditional-density model, it
uses an arm-specific location-shift approximation.
The complete residual pool is averaged deterministically for every
query, without Monte Carlo mediator sampling.
Only mediator residual construction is cross-fitted; the treatment
and outcome models use the full context.

\paragraph{Instrumental variable (IV).}
For IV, the modular conditional Wald estimator~\citep{wald1940fitting, angrist1995identification, wang2018bounded} uses supervised models
to estimate the conditional reduced-form outcome and treatment
responses to the instrument.
In the reported Wald configuration, these predictive components use
TabPFN-v3.5.
Algorithm~\ref{alg:conditional-wald} gives the complete procedure.

\begin{algorithm}[t]
\caption{Conditional Wald estimator}
\label{alg:conditional-wald}
\begin{algorithmic}[1]
\Require Context data $\{(X_i,I_i,T_i,Y_i)\}_{i=1}^{n}$, query covariate $x$
\Ensure Conditional Wald estimate $\widehat\tau_{\mathrm{IV}}(x)$
\State Fit outcome model $\widehat\mu_Y(j,x)\approx \mathbb{E}[Y\mid I=j,X=x]$
\State Fit treatment model $\widehat\mu_T(j,x)\approx P(T=1\mid I=j,X=x)$
\For{$j\in\{0,1\}$}
    \State Evaluate $\widehat\mu_Y(j,x)$ and $\widehat\mu_T(j,x)$
\EndFor
\State Compute
$
\widehat\Delta_Y(x)
=
\widehat\mu_Y(1,x)-\widehat\mu_Y(0,x), \quad 
\widehat\Delta_T(x)
=
\widehat\mu_T(1,x)-\widehat\mu_T(0,x)
$
\State \Return
$
\widehat\tau_{\mathrm{IV}}(x)
=
\frac{\widehat\Delta_Y(x)}{\widehat\Delta_T(x)}
$
\end{algorithmic}
\end{algorithm}

Both predictive models are fitted once on the full context and are
not cross-fitted.
We preserve the sign of the first-stage contrast and apply no
denominator floor, additive epsilon, fallback prediction, or
final-effect clipping.
If $|\widehat\Delta_T(x)|\leq 10^{-12}$, the estimate remains
undefined and the run fails the subsequent finiteness check.
The benchmark's within-$X$ gain restriction makes the conditional
Wald estimand coincide with $\tau_0(x)$.

\paragraph{Proximal (PX).}
We implement a common P-Learner \citep{sverdrup2023proximal} with
ExtraTrees, neural networks, or TabPFN-v3.5 as the predictive
backbone.
The backbone is changed jointly for all nuisance models and the final
pseudo-outcome regression, while the cross-fitting layout, bridge
construction, and proximal score remain fixed.
Algorithm~\ref{alg:p-learner} summarizes the procedure.

\begin{algorithm}[t]
\caption{Proximal P-Learner}
\label{alg:p-learner}
\begin{algorithmic}[1]
\Require Context data $\{(X_i,Z_{p,i},W_{p,i},T_i,Y_i)\}_{i=1}^{n}$, query covariate $x$
\Ensure Proximal estimate $\widehat\tau_{\mathrm{PX}}(x)$
\State Split the context into five treatment-stratified folds
\For{fold $k=1,\dots,5$}
    \State Fit
    $e_w(x)\approx P(T=1\mid X=x,W_p=w)$
    on the other four folds
    \For{$a\in\{0,1\}$}
        \State Fit $\mu_{az}(x)\approx \mathbb{E}[Y\mid X=x,Z_p=z,T=a]$
        \State Fit $p^W_{az}(x)\approx P(W_p=1\mid X=x,Z_p=z,T=a)$
        \State Fit $p^Z_{aw}(x)\approx P(Z_p=1\mid X=x,W_p=w,T=a)$
        \State Solve the regularized binary outcome bridge
        $h_a(w,x)$
        \State Solve the regularized binary treatment bridge
        $q_a(z,x)$
    \EndFor
    \For{held-out observation $i$ in fold $k$}
        \State Clip the factual treatment-bridge evaluation:
        \[
        \widetilde q_{T_i}(Z_{p,i},X_i)
        =
        \operatorname{clip}_{[-25,25]}
        \bigl(q_{T_i}(Z_{p,i},X_i)\bigr)
        \]
        \State Compute the proximal pseudo-outcome
        \[
        \Gamma_i
        =
        (2T_i-1)\,
        \widetilde q_{T_i}(Z_{p,i},X_i)
        \{Y_i-h_{T_i}(W_{p,i},X_i)\}
        +
        h_1(W_{p,i},X_i)-h_0(W_{p,i},X_i)
        \]
    \EndFor
\EndFor
\State Fit the final regression $
\widehat\tau_{\mathrm{PX}}(x)\approx \mathbb{E}[\Gamma\mid X=x]$
using $\{(X_i,\Gamma_i)\}_{i=1}^{n}$
\State \Return $\widehat\tau_{\mathrm{PX}}(x)$
\end{algorithmic}
\end{algorithm}

Because both proxies are binary, we solve the outcome and treatment
bridges using explicit ridge-regularized two-point equations with
ridge parameter $0.01$ rather than fitting separate bridge models
directly.
All nuisance quantities entering $\Gamma_i$ are out of fold.

Across the ExtraTrees, neural-network, and TabPFN-v3.5 variants, the
same bridge equations and pseudo-outcome are retained.
Only the predictive backbone used for the nuisance estimates and the
final regression of $\Gamma$ on $X$ changes. The query input therefore contains $X$ only. The proxies contribute
through the context-level bridge and pseudo-outcome construction.

\subsubsection{Additional point-estimation baselines}
\label{app:additional_point_baselines}

ForestDRIV combines the doubly robust instrumental-variable
(DRIV) procedure \citep{syrgkanis2019machine} with a regression
forest as the final effect model. The procedure constructs a residual-corrected regression target
from estimated nuisance functions and a preliminary effect model,
then regresses this target on the effect-modifying covariates. KIV \citep{singh2019kernel} is a nonparametric IV estimator based
on two-stage kernel ridge regression.
It estimates a conditional mean embedding of the
structural inputs given the instruments and then uses
these embeddings to estimate the structural outcome response.

Causal Tree \citep{athey2016recursive} recursively partitions
the covariate space to capture treatment-effect heterogeneity.
It estimates a treatment effect within each terminal node and
assigns that estimate to query observations falling in the node,
yielding a piecewise-constant effect function. Generalized random forests (GRF) \citep{athey2019generalized}
estimate quantities defined by local moment equations using
adaptive weights learned by a forest.
The causal-forest specialization uses splits designed to
capture treatment-effect heterogeneity and estimates conditional
effects within the resulting forest-weighted neighborhoods.

The treatment-agnostic representation network (TARNet)
\citep{shalit2017estimating} learns a shared covariate
representation with separate outcome heads for the treated
and control groups.
Training uses observed outcomes through the corresponding
treatment head without an explicit representation-balancing
penalty. CFRNet \citep{shalit2017estimating} augments the shared-representation,
treatment-specific-head architecture with a penalty on the
distributional discrepancy between treated and control
representations.
Its objective combines factual outcome prediction with this
balancing term.
As with TARNet, treatment effects are predicted by contrasting
the two outcome heads at the same query covariates.

\subsubsection{Models for Partial Identification}
\label{app:partial_id_baseline}
We evaluate bound estimators on the matched binary companion worlds
in Appendix~\ref{app:partial_id_scm}.
For Manski and IV bounds, multinomial logistic regression,
XGBoost \citep{chen2016xgboost}, and frozen TabPFN-v3.5
\citep{jager2026tabpfn} estimate $P(T,Y\mid X)$ and
$P(T,Y\mid X,I)$, respectively.
The Manski formula \citep{manski2003partial} and the IV
response-type linear program \citep{balke1997bounds} are held fixed
across backbones, with treatment monotonicity imposed in the latter.
We apply a common feasibility projection to infeasible estimated
IV cells.
For sensitivity bounds, we evaluate CSA-PFN (MSM)
\citep{javurek2026amortizing}, B-Learner \citep{oprescu2023b},
and NeuralCSA \citep{frauen2024neural} at
$\Gamma\in\{1,1.25,1.5,2,3,5\}$.
In our implementation, CSA-PFN uses its frozen checkpoint with a
context-fitted 50-to-10 PCA adapter, while B-Learner uses five-fold
cross-fitting with random forests. For the B-Learner variants, ``all'' uses the indicated predictive backbone for all learned components, whereas
``final'' replaces only the final bound regressions with
TabPFN and retains RF-based nuisance estimators.
NeuralCSA is trained separately in each world, applying its
continuous-density flow to binary outcomes.
All estimators use only context data for fitting and are evaluated
against the corresponding SCM oracle bounds on shared query units.

\subsection{Semi-Synthetic Evaluation Settings}
\label{app:semi-synthetic}

For ACIC 2016~\citep{dorie2019automated}, we select 40 parameter
settings by rounding 40 equally spaced indices from 1 to 77,
using simulation 1 for each setting.
For the LaLonde-PSID and LaLonde-CPS
datasets~\citep{lalonde1986evaluating,dehejia1999causal,dehejia2002propensity}
generated using RealCause~\citep{neal2020realcause},
we use official samples 0--39.
From each realization, we select 1,024 context units and
100 disjoint query units, with identical context and query units
across estimators.
Within each RealCause dataset, covariate rows and split indices
are also preserved across realizations when the archived samples
support exact pairing.
ACIC retains its original 58-column schema, with the three
string-valued columns encoded ordinally using sorted category
levels, whereas RealCause uses its eight native numeric covariates.
The CATE targets are conditional-mean contrasts from the official
ACIC generator or the pre-trained RealCause outcome generator.

\subsection{Real-World Evaluation Settings}
\label{app:real-world}

The control--treatment contrasts are defined as
AZT monotherapy versus AZT+ddI combination therapy in
ACTG 175~\citep{hammer1996trial},
group 0 versus group 6 among records satisfying
\texttt{dem\_inel=1} and \texttt{revasamp=1} in
Pennsylvania~\citep{corson1992pennsylvania},
and the original control group versus the job-search incentive
group in Illinois~\citep{woodbury1987bonuses}.
The respective negative-control outcomes are
\texttt{cd40}, \texttt{wages}, and \texttt{avprearn}.
Context sampling is treatment-stratified to approximately
preserve the eligible sample's treatment proportion, with
768, 1,024, and 2,048 context units for ACTG 175,
Pennsylvania, and Illinois, respectively.
Each split contains 256 query units sampled without replacement
from the remaining records, and all estimators share the same
context and query units.
All pre-processing steps are fitted using the context data only.
Numeric features undergo median imputation, categorical features
undergo mode imputation and one-hot encoding, and outcomes are
standardized using the context mean and population standard deviation.
We compute metrics relative to the null target $\tau(x)=0$
on the standardized outcome scale and average them over the 40 splits.
These repetitions assess split sensitivity within a fixed dataset,
rather than variation across independent trials.

Although the original trials contain multiple treatment arms, we form
each benchmark contrast by selecting two randomized arms and designating
one as control and the other as treatment.
Let $A\in\mathcal A$ denote the original treatment assignment.
For selected arms $a_0,a_1\in\mathcal A$, define
$S=\mathbbm{1}\{A\in\{a_0,a_1\}\}$ and $T=\mathbbm{1}\{A=a_1\}$.
Since the negative-control outcome is measured before treatment,
it cannot be affected by treatment assignment~\citep{arnold2016negative}.
In the potential-outcome formulation~\citep{ashby2025negative},
\[
Y^{\mathrm{pre}}(a_0)
=
Y^{\mathrm{pre}}(a_1)
=
Y^{\mathrm{pre}},
\]
and therefore
\[
\tau(x)
=
\mathbb E\!\left[
Y^{\mathrm{pre}}(a_1)-Y^{\mathrm{pre}}(a_0)
\mid X=x,S=1
\right]
=0.
\]
Moreover, randomization of the original arm assignment implies
\[
A \perp
\{Y^{\mathrm{pre}}(a):a\in\mathcal A\}
\mid X,
\]
and hence, after restricting to the selected pair,
\[
T \perp
\bigl\{
Y^{\mathrm{pre}}(a_0),
Y^{\mathrm{pre}}(a_1)
\bigr\}
\mid X,S=1.
\]

\subsection{IHDP Full versus IHDP-HC}
\label{app:ihdp_full_hidden}

We adopt the Infant Health and Development Program (IHDP) benchmark
to evaluate treatment-effect estimation under confounder omission.
Following the hidden-confounding construction of
\citet{jesson2021quantifying} and its Quince implementation, we
combine empirical covariates and treatment assignments with
synthetically generated outcomes. The resulting dataset contains
747 units with 25 covariates. The Full view retains all covariates,
including $U=\texttt{b.marr}$, the mother's marital status at
childbirth. The IHDP-HC view removes only $U$ from the model inputs,
leaving 24 covariates. The omitted variable remains part of the
outcome-generating process and can also modify treatment effects. Unlike the main synthetic benchmark, the IHDP-HC comparison also removes an effect modifier. Its error increase therefore combines confounding-related degradation with loss of information about effect heterogeneity.

We evaluate 100 paired trials using the Quince data preprocessing
and splitting procedure. Each trial contains 470 training,
202 validation, and 75 test units. For the main comparison, the training and validation subsets are pooled into a context of
672 units, with the 75 test units used as queries. Within each trial,
Full and Hidden share the same unit identities, treatment assignments,
factual and potential outcomes, split, and oracle effect values.
Both views are evaluated against the same
$\tau_{\mathrm{true}}(X,U)$ values, without redefining the target
after removing $U$. This paired comparison serves as a
semi-synthetic stress test of point estimation when a confounder
is withheld from the estimator.

\section{Ablation Studies}
\subsection{Hidden Confounder Track}
\label{sec:hidden}
\begin{figure*}[t]
    \centering
    \setlength{\abovecaptionskip}{2pt}
    \setlength{\parskip}{0pt}
    \includegraphics[width=\textwidth]{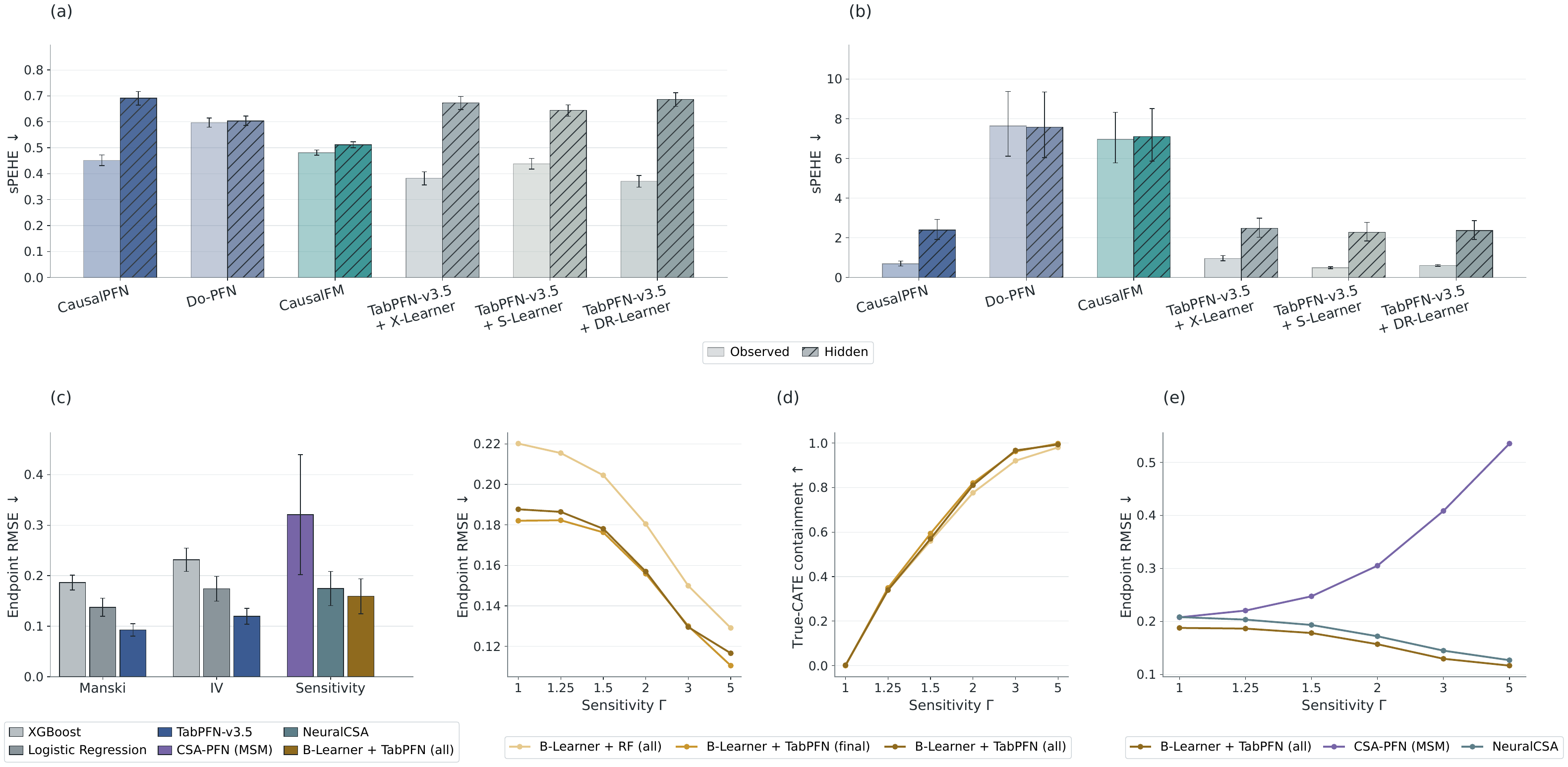}\par
    \caption{
    \textbf{Hidden-confounding and partial-identification evaluation.}
    (a) Synthetic paired comparison of point-estimation error when
    the confounder is observed (BD) or unobserved (HC).
    (b) Full versus Hidden comparisons on
    IHDP-HC. (c) Endpoint RMSE for representative partially identified
    settings, including bounded-outcome, IV-based, and sensitivity-model
    leaves. The sensitivity comparison in panel (c) is evaluated at
$\Gamma=2$. (d) Performance of B-Learner variants across sensitivity levels
    $\Gamma$.
    (e) Endpoint RMSE of sensitivity-aware estimators across
    $\Gamma$.
    }
    \label{fig:hidden_confounder_track}
\end{figure*}

Figure~\ref{fig:hidden_confounder_track} examines point-estimator behavior under unobserved confounding and the accuracy of estimated partial-identification bounds.
The synthetic HC view retains only $(X,T,Y)$, without the confounder
or auxiliary identifying variables.
Point predictions in this HC view are evaluated without a point-identification claim.

\paragraph{Confounder omission affects models differently.}
Panels (a) and (b) compare point-estimation performance with the
designated confounder observed or unobserved on synthetic and IHDP-HC data, respectively.
CausalPFN and all three TabPFN-based modular approaches exhibit pronounced increases in sPEHE after confounder omission in both settings, whereas Do-PFN and CausalFM show smaller changes. 

\paragraph{TabPFN improves bound-estimation accuracy.}
Panel (c) evaluates three representative partial-identification settings.
TabPFN-v3.5 achieves the lowest endpoint RMSE among the compared
estimators in the Manski and IV settings, while
B-Learner + TabPFN (all) achieves the lowest error in the
sensitivity setting.
Panel (d) further compares RF- and TabPFN-based B-Learner variants.
Both TabPFN configurations obtain lower endpoint RMSE than the
RF-based configuration at every evaluated sensitivity level
$\Gamma$.
Their target-CATE containment curves increase with
$\Gamma$.
These results distinguish accurate recovery of the reference
bounds from inclusion of the oracle CATE and support the use of
TabPFN within bound-estimation procedures.

\paragraph{Sensitivity-level dependence varies across estimators.}
In panel (e), the endpoint RMSE of CSA-PFN (MSM) increases
monotonically over the evaluated sensitivity grid.
In contrast, B-Learner + TabPFN (all) and NeuralCSA exhibit
decreasing endpoint errors, with the TabPFN-based B-Learner
achieving the lowest error at every evaluated $\Gamma$.
Its advantage over CSA-PFN becomes larger as
the maintained sensitivity restriction is relaxed.
This contrast highlights the importance of evaluating
sensitivity-bound estimators across multiple sensitivity levels.

\subsection{Stress Test for Synthetic Data}

\paragraph{Regime-specific stress.}
Table~\ref{tab:regime-specific-stress} highlights both the strengths
and limitations of modular estimation under regime-specific stress.
In BD and FD, the TabPFN-based DR-Learner and front-door plug-in,
respectively, retain the lowest mean $\mathrm{sPEHE}^{2}$ despite
weaker treatment overlap and reduced mediator support.
All evaluated plug-in estimators in the FD comparison outperform the CFM configurations throughout the stress sweep.
However, this advantage does not extend to IV and PX.
Under weaker instruments, the TabPFN-based Wald estimator and
ForestDRIV exhibit sharply increasing errors and between-world
variability, while CausalFM maintains the lowest mean error.
In PX, worsening bridge conditioning erodes the initial advantage
of the TabPFN-based P-Learner, with CausalFM becoming the
lowest-error configuration under the Severe setting.
These contrasting results show that the empirical benefits of
modular estimation depend not only on the identification regime
but also on the conditions under which its estimation procedure
is applied.

\begin{table*}[t]
\centering
\caption{
\textbf{Regime-specific stress tests across four identification regimes.}
We report $\mathrm{sPEHE}^{2}$ ($\downarrow$) as the mean $\pm$
standard deviation across 40 SCM worlds.
Default, Mild, Moderate, and Severe denote increasing levels of regime-specific stress. These levels are not calibrated across regimes.
\textbf{Bold} means indicate the lowest mean error within each regime
and severity level.
}
\label{tab:regime-specific-stress}
\begingroup
\fontsize{8.5}{10.5}\selectfont
\setlength{\tabcolsep}{3pt}
\renewcommand{\arraystretch}{1.12}
\begin{tabular*}{\textwidth}{@{\extracolsep{\fill}}lcccc@{}}
\toprule
& \multicolumn{4}{c}{Stress severity} \\
\cmidrule(l){2-5}
Method & Default & Mild & Moderate & Severe \\
\midrule
\multicolumn{5}{@{}l}{\textbf{Back-door (BD): Overlap}} \\
\addlinespace[2pt]
CausalPFN
& $0.208\,{\scriptstyle \pm\,0.062}$
& $0.233\,{\scriptstyle \pm\,0.068}$
& $0.253\,{\scriptstyle \pm\,0.074}$
& $0.291\,{\scriptstyle \pm\,0.084}$ \\
Do-PFN
& $0.360\,{\scriptstyle \pm\,0.071}$
& $0.338\,{\scriptstyle \pm\,0.066}$
& $0.322\,{\scriptstyle \pm\,0.064}$
& $0.310\,{\scriptstyle \pm\,0.061}$ \\
CausalFM
& $0.233\,{\scriptstyle \pm\,0.033}$
& $0.232\,{\scriptstyle \pm\,0.033}$
& $0.234\,{\scriptstyle \pm\,0.034}$
& $0.235\,{\scriptstyle \pm\,0.034}$ \\
\addlinespace[3pt]
TabPFN-v3.5 + X-Learner
& $0.153\,{\scriptstyle \pm\,0.060}$
& $0.166\,{\scriptstyle \pm\,0.066}$
& $0.182\,{\scriptstyle \pm\,0.068}$
& $0.212\,{\scriptstyle \pm\,0.078}$ \\
TabPFN-v3.5 + S-Learner
& $0.197\,{\scriptstyle \pm\,0.057}$
& $0.199\,{\scriptstyle \pm\,0.058}$
& $0.205\,{\scriptstyle \pm\,0.054}$
& $0.209\,{\scriptstyle \pm\,0.054}$ \\
TabPFN-v3.5 + DR-Learner
& $\mathbf{0.143}\,{\scriptstyle \pm\,0.054}$
& $\mathbf{0.154}\,{\scriptstyle \pm\,0.060}$
& $\mathbf{0.171}\,{\scriptstyle \pm\,0.063}$
& $\mathbf{0.202}\,{\scriptstyle \pm\,0.070}$ \\
\midrule
\multicolumn{5}{@{}l}{\textbf{Front-door (FD): Mediator support}} \\
\addlinespace[2pt]
CausalPFN 
& $0.485\,{\scriptstyle \pm\,0.117}$
& $0.480\,{\scriptstyle \pm\,0.115}$
& $0.476\,{\scriptstyle \pm\,0.114}$
& $0.473\,{\scriptstyle \pm\,0.113}$ \\
Do-PFN 
& $0.368\,{\scriptstyle \pm\,0.074}$
& $0.365\,{\scriptstyle \pm\,0.074}$
& $0.364\,{\scriptstyle \pm\,0.075}$
& $0.364\,{\scriptstyle \pm\,0.075}$ \\
CausalFM
& $0.551\,{\scriptstyle \pm\,0.073}$
& $0.550\,{\scriptstyle \pm\,0.073}$
& $0.549\,{\scriptstyle \pm\,0.073}$
& $0.549\,{\scriptstyle \pm\,0.073}$ \\
\addlinespace[3pt]
NN + FD Plug-in
& $0.227\,{\scriptstyle \pm\,0.031}$
& $0.230\,{\scriptstyle \pm\,0.033}$
& $0.232\,{\scriptstyle \pm\,0.033}$
& $0.235\,{\scriptstyle \pm\,0.032}$ \\
XGBoost + FD Plug-in
& $0.233\,{\scriptstyle \pm\,0.037}$
& $0.227\,{\scriptstyle \pm\,0.041}$
& $0.234\,{\scriptstyle \pm\,0.042}$
& $0.251\,{\scriptstyle \pm\,0.054}$ \\
TabPFN-v3.5 + FD Plug-in
& $\mathbf{0.154}\,{\scriptstyle \pm\,0.057}$
& $\mathbf{0.181}\,{\scriptstyle \pm\,0.051}$
& $\mathbf{0.198}\,{\scriptstyle \pm\,0.049}$
& $\mathbf{0.212}\,{\scriptstyle \pm\,0.050}$ \\
\midrule
\multicolumn{5}{@{}l}{\textbf{Instrumental variable (IV): Instrument strength}} \\
\addlinespace[2pt]
CausalPFN
& $0.518\,{\scriptstyle \pm\,0.127}$
& $0.528\,{\scriptstyle \pm\,0.124}$
& $0.542\,{\scriptstyle \pm\,0.120}$
& $0.540\,{\scriptstyle \pm\,0.118}$ \\
Do-PFN
& $0.371\,{\scriptstyle \pm\,0.073}$
& $0.369\,{\scriptstyle \pm\,0.072}$
& $0.366\,{\scriptstyle \pm\,0.071}$
& $0.365\,{\scriptstyle \pm\,0.071}$ \\
CausalFM
& $\mathbf{0.258}\,{\scriptstyle \pm\,0.041}$
& $\mathbf{0.256}\,{\scriptstyle \pm\,0.039}$
& $\mathbf{0.253}\,{\scriptstyle \pm\,0.039}$
& $\mathbf{0.253}\,{\scriptstyle \pm\,0.039}$ \\
\addlinespace[3pt]
ForestDRIV
& $0.460\,{\scriptstyle \pm\,0.264}$
& $1.770\,{\scriptstyle \pm\,4.671}$
& $26.980\,{\scriptstyle \pm\,43.796}$
& $111.043\,{\scriptstyle \pm\,156.843}$ \\
KIV
& $0.420\,{\scriptstyle \pm\,0.187}$
& $0.457\,{\scriptstyle \pm\,0.189}$
& $0.468\,{\scriptstyle \pm\,0.187}$
& $0.491\,{\scriptstyle \pm\,0.209}$ \\
TabPFN-v3.5 + Wald
& $0.336\,{\scriptstyle \pm\,0.105}$
& $0.389\,{\scriptstyle \pm\,0.174}$
& $1.145\,{\scriptstyle \pm\,2.578}$
& $1.94\times10^{7}\,{\scriptstyle \pm\,1.22\times10^{8}}$ \\
\midrule
\multicolumn{5}{@{}l}{\textbf{Proximal (PX): Bridge conditioning}} \\
\addlinespace[2pt]
CausalPFN
& $0.243\,{\scriptstyle \pm\,0.056}$
& $0.287\,{\scriptstyle \pm\,0.069}$
& $0.352\,{\scriptstyle \pm\,0.080}$
& $0.445\,{\scriptstyle \pm\,0.101}$ \\
Do-PFN
& $0.347\,{\scriptstyle \pm\,0.066}$
& $0.351\,{\scriptstyle \pm\,0.067}$
& $0.356\,{\scriptstyle \pm\,0.070}$
& $0.362\,{\scriptstyle \pm\,0.071}$ \\
CausalFM
& $0.240\,{\scriptstyle \pm\,0.033}$
& $0.250\,{\scriptstyle \pm\,0.034}$
& $0.269\,{\scriptstyle \pm\,0.035}$
& $\mathbf{0.301}\,{\scriptstyle \pm\,0.043}$ \\
\addlinespace[3pt]
ExtraTrees + P-Learner
& $0.233\,{\scriptstyle \pm\,0.049}$
& $0.284\,{\scriptstyle \pm\,0.065}$
& $0.318\,{\scriptstyle \pm\,0.062}$
& $0.441\,{\scriptstyle \pm\,0.094}$ \\
NN + P-Learner
& $0.662\,{\scriptstyle \pm\,1.199}$
& $0.471\,{\scriptstyle \pm\,0.247}$
& $0.365\,{\scriptstyle \pm\,0.121}$
& $0.536\,{\scriptstyle \pm\,0.213}$ \\
TabPFN-v3.5 + P-Learner
& $\mathbf{0.181}\,{\scriptstyle \pm\,0.054}$
& $\mathbf{0.193}\,{\scriptstyle \pm\,0.056}$
& $\mathbf{0.217}\,{\scriptstyle \pm\,0.052}$
& $0.442\,{\scriptstyle \pm\,0.108}$ \\
\bottomrule
\end{tabular*}
\endgroup
\end{table*}

\paragraph{Context length.}
Table~\ref{tab:context-length-stress} shows that smaller contexts
generally reduce estimation accuracy and can change which
configuration performs best.
While the TabPFN-based FD plug-in and CausalFM in IV retain
the lowest mean errors throughout the sweep, the leading
configurations in BD and PX change at the shortest context.
The extent of degradation also differs substantially across
estimators.
CausalPFN's mean error more than doubles in BD and PX, and
ForestDRIV exhibits pronounced increases in both error and
between-world variability in IV.
By contrast, Do-PFN's errors change relatively little across
the four regimes.
Such limited sensitivity should not, however, be interpreted
as evidence of accurate estimation: CausalFM's FD error remains
nearly unchanged yet consistently exceeds that of every FD
plug-in estimator.
Thus, context-length evaluation reveals both the dependence of
relative performance on data availability and the distinction
between maintaining a similar error and maintaining a low error.

\begin{table*}[t]
\centering
\caption{
\textbf{Context-length stress test across four identification regimes.}
We report $\mathrm{sPEHE}^{2}$ ($\downarrow$) as the mean $\pm$
standard deviation across 40 SCM worlds.
Context lengths of 1,024, 512, 256, and 160 correspond to the
Default, Mild, Moderate, and Severe settings, respectively.
\textbf{Bold} means indicate the lowest mean error within each regime
and context length.
}
\label{tab:context-length-stress}
\begingroup
\fontsize{8.5}{10.5}\selectfont
\setlength{\tabcolsep}{3pt}
\renewcommand{\arraystretch}{1.12}
\begin{tabular*}{\textwidth}{@{\extracolsep{\fill}}lcccc@{}}
\toprule
& \multicolumn{4}{c}{Context length $n_{\mathrm{ctx}}$} \\
\cmidrule(l){2-5}
Method & Default & Mild & Moderate & Severe \\
\midrule
\multicolumn{5}{@{}l}{\textbf{Back-door (BD)}} \\
\addlinespace[2pt]
CausalPFN
& $0.208\,{\scriptstyle \pm\,0.062}$
& $0.249\,{\scriptstyle \pm\,0.063}$
& $0.327\,{\scriptstyle \pm\,0.085}$
& $0.451\,{\scriptstyle \pm\,0.134}$ \\
Do-PFN
& $0.360\,{\scriptstyle \pm\,0.071}$
& $0.363\,{\scriptstyle \pm\,0.071}$
& $0.370\,{\scriptstyle \pm\,0.079}$
& $0.383\,{\scriptstyle \pm\,0.082}$ \\
CausalFM
& $0.233\,{\scriptstyle \pm\,0.033}$
& $0.240\,{\scriptstyle \pm\,0.035}$
& $0.253\,{\scriptstyle \pm\,0.044}$
& $0.273\,{\scriptstyle \pm\,0.055}$ \\
\addlinespace[3pt]
TabPFN-v3.5 + X-Learner
& $0.153\,{\scriptstyle \pm\,0.060}$
& $0.202\,{\scriptstyle \pm\,0.056}$
& $0.261\,{\scriptstyle \pm\,0.069}$
& $0.308\,{\scriptstyle \pm\,0.089}$ \\
TabPFN-v3.5 + S-Learner
& $0.197\,{\scriptstyle \pm\,0.057}$
& $0.227\,{\scriptstyle \pm\,0.039}$
& $0.246\,{\scriptstyle \pm\,0.039}$
& $\mathbf{0.268}\,{\scriptstyle \pm\,0.053}$ \\
TabPFN-v3.5 + DR-Learner
& $\mathbf{0.143}\,{\scriptstyle \pm\,0.054}$
& $\mathbf{0.197}\,{\scriptstyle \pm\,0.049}$
& $\mathbf{0.244}\,{\scriptstyle \pm\,0.068}$
& $0.289\,{\scriptstyle \pm\,0.095}$ \\
\midrule
\multicolumn{5}{@{}l}{\textbf{Front-door (FD)}} \\
\addlinespace[2pt]
CausalPFN
& $0.485\,{\scriptstyle \pm\,0.117}$
& $0.542\,{\scriptstyle \pm\,0.116}$
& $0.580\,{\scriptstyle \pm\,0.159}$
& $0.691\,{\scriptstyle \pm\,0.225}$ \\
Do-PFN
& $0.368\,{\scriptstyle \pm\,0.074}$
& $0.372\,{\scriptstyle \pm\,0.072}$
& $0.381\,{\scriptstyle \pm\,0.082}$
& $0.391\,{\scriptstyle \pm\,0.084}$ \\
CausalFM
& $0.551\,{\scriptstyle \pm\,0.073}$
& $0.551\,{\scriptstyle \pm\,0.074}$
& $0.552\,{\scriptstyle \pm\,0.074}$
& $0.552\,{\scriptstyle \pm\,0.074}$ \\
\addlinespace[3pt]
NN + FD Plug-in
& $0.227\,{\scriptstyle \pm\,0.031}$
& $0.240\,{\scriptstyle \pm\,0.039}$
& $0.282\,{\scriptstyle \pm\,0.067}$
& $0.286\,{\scriptstyle \pm\,0.066}$ \\
XGBoost + FD Plug-in
& $0.233\,{\scriptstyle \pm\,0.037}$
& $0.271\,{\scriptstyle \pm\,0.043}$
& $0.299\,{\scriptstyle \pm\,0.053}$
& $0.313\,{\scriptstyle \pm\,0.057}$ \\
TabPFN-v3.5 + FD Plug-in
& $\mathbf{0.154}\,{\scriptstyle \pm\,0.057}$
& $\mathbf{0.212}\,{\scriptstyle \pm\,0.050}$
& $\mathbf{0.242}\,{\scriptstyle \pm\,0.047}$
& $\mathbf{0.260}\,{\scriptstyle \pm\,0.048}$ \\
\midrule
\multicolumn{5}{@{}l}{\textbf{Instrumental variable (IV)}} \\
\addlinespace[2pt]
CausalPFN
& $0.518\,{\scriptstyle \pm\,0.127}$
& $0.570\,{\scriptstyle \pm\,0.127}$
& $0.593\,{\scriptstyle \pm\,0.168}$
& $0.691\,{\scriptstyle \pm\,0.224}$ \\
Do-PFN
& $0.371\,{\scriptstyle \pm\,0.073}$
& $0.374\,{\scriptstyle \pm\,0.072}$
& $0.385\,{\scriptstyle \pm\,0.081}$
& $0.394\,{\scriptstyle \pm\,0.083}$ \\
CausalFM
& $\mathbf{0.258}\,{\scriptstyle \pm\,0.041}$
& $\mathbf{0.262}\,{\scriptstyle \pm\,0.039}$
& $\mathbf{0.273}\,{\scriptstyle \pm\,0.043}$
& $\mathbf{0.288}\,{\scriptstyle \pm\,0.055}$ \\
\addlinespace[3pt]
ForestDRIV
& $0.460\,{\scriptstyle \pm\,0.264}$
& $1.561\,{\scriptstyle \pm\,3.460}$
& $11.620\,{\scriptstyle \pm\,38.073}$
& $29.283\,{\scriptstyle \pm\,58.180}$ \\
KIV
& $0.420\,{\scriptstyle \pm\,0.187}$
& $0.432\,{\scriptstyle \pm\,0.231}$
& $0.443\,{\scriptstyle \pm\,0.192}$
& $0.461\,{\scriptstyle \pm\,0.179}$ \\
TabPFN-v3.5 + Wald
& $0.336\,{\scriptstyle \pm\,0.105}$
& $0.372\,{\scriptstyle \pm\,0.159}$
& $0.402\,{\scriptstyle \pm\,0.095}$
& $0.477\,{\scriptstyle \pm\,0.331}$ \\
\midrule
\multicolumn{5}{@{}l}{\textbf{Proximal (PX)}} \\
\addlinespace[2pt]
CausalPFN
& $0.243\,{\scriptstyle \pm\,0.056}$
& $0.296\,{\scriptstyle \pm\,0.059}$
& $0.354\,{\scriptstyle \pm\,0.087}$
& $0.505\,{\scriptstyle \pm\,0.149}$ \\
Do-PFN
& $0.347\,{\scriptstyle \pm\,0.066}$
& $0.351\,{\scriptstyle \pm\,0.066}$
& $0.360\,{\scriptstyle \pm\,0.074}$
& $0.373\,{\scriptstyle \pm\,0.078}$ \\
CausalFM
& $0.240\,{\scriptstyle \pm\,0.033}$
& $0.251\,{\scriptstyle \pm\,0.039}$
& $0.267\,{\scriptstyle \pm\,0.051}$
& $\mathbf{0.288}\,{\scriptstyle \pm\,0.057}$ \\
\addlinespace[3pt]
ExtraTrees + P-Learner
& $0.233\,{\scriptstyle \pm\,0.049}$
& $0.300\,{\scriptstyle \pm\,0.069}$
& $0.383\,{\scriptstyle \pm\,0.101}$
& $0.516\,{\scriptstyle \pm\,0.275}$ \\
NN + P-Learner
& $0.662\,{\scriptstyle \pm\,1.199}$
& $0.699\,{\scriptstyle \pm\,0.576}$
& $0.889\,{\scriptstyle \pm\,1.914}$
& $0.851\,{\scriptstyle \pm\,0.911}$ \\
TabPFN-v3.5 + P-Learner
& $\mathbf{0.181}\,{\scriptstyle \pm\,0.054}$
& $\mathbf{0.222}\,{\scriptstyle \pm\,0.048}$
& $\mathbf{0.258}\,{\scriptstyle \pm\,0.080}$
& $0.394\,{\scriptstyle \pm\,0.382}$ \\
\bottomrule
\end{tabular*}
\endgroup
\end{table*}

\subsection{Identification Beyond Additional Inputs}
\begin{figure*}[t]
    \centering
    \setlength{\abovecaptionskip}{2pt}
    \setlength{\parskip}{0pt}
    \includegraphics[width=\textwidth]{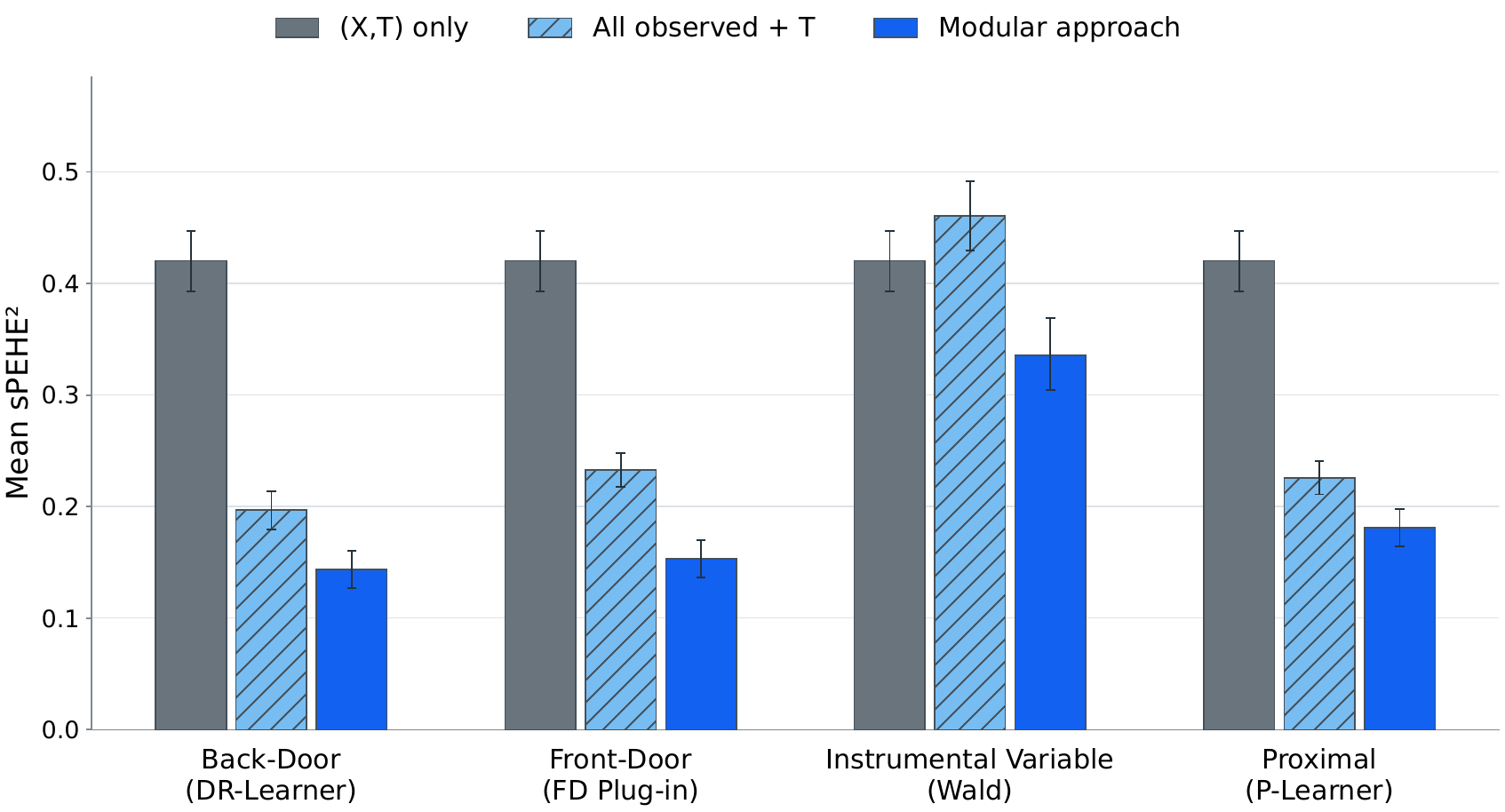}\par
    \caption{
    \textbf{Input and identification ablation for the modular approach.}
    We compare three TabPFN configurations: $(X,T)$ only, all variables
    available in each observational view provided as ordinary predictive
    features, and the corresponding regime-specific modular estimator.
    The first two configurations provide no annotations or information
    about the causal roles of the input variables.
    }
    \label{fig:TabPFN_input}
\end{figure*}

Figure~\ref{fig:TabPFN_input} compares three TabPFN configurations to distinguish the effect of additional observed variables from that of explicit regime-specific causal estimation.
The first configuration uses TabPFN to predict $Y$ from $(X,T)$
alone, excluding all auxiliary variables and providing a common reference across views.
The second configuration additionally supplies all auxiliary variables available in the corresponding observational view as ordinary predictive features. Neither configuration provides annotations about the causal roles of these variables or applies a regime-specific identification procedure. The third integrates TabPFN into the corresponding modular estimator.

The modular configuration achieves the lowest mean sPEHE$^2$
across all four evaluated views.
Adding auxiliary variables as ordinary features improves
performance over the $(X,T)$-only reference in the back-door, front-door, and proximal views, but increases error in the IV view. Even when auxiliary variables improve performance, the corresponding predictive configurations still have higher errors than the modular estimators. These results support combining TabPFN's predictive capabilities
with regime-specific identification.

\subsection{Effect of the Predictive Backbone in Modular Approaches}

We examine how the predictive backbone affects the modular
estimators while keeping the regime-specific identification
procedure fixed. For each regime, we replace the predictive model
used within the corresponding X-Learner, front-door plug-in, Wald,
or P-Learner pipeline with alternative tabular predictors.

\begin{figure*}[t]
    \centering
    \setlength{\abovecaptionskip}{2pt}
    \setlength{\parskip}{0pt}
    \includegraphics[width=\textwidth]{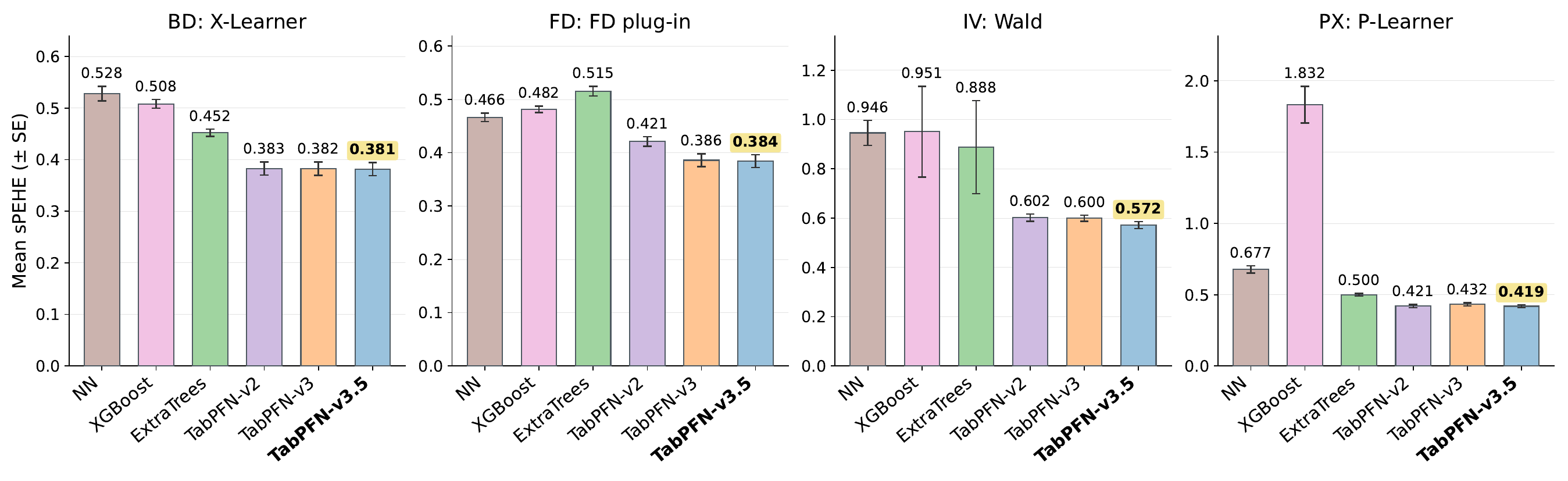}\par
    \caption{
    \textbf{Effect of the predictive backbone in modular approaches.}
    We replace the predictive model in modular approaches while keeping the regime-specific
    identification procedure fixed within each panel. 
    }
    \label{fig:modular_backend}
\end{figure*}

Figure~\ref{fig:modular_backend} shows that the choice of predictive backbone can substantially affect causal estimation performance, even when the identification procedure is unchanged.
The TabPFN variants generally achieve lower mean sPEHE than the
conventional predictive models across the four regimes.
Performance is relatively stable across versions of TabPFN. TabPFN-v3.5~\citep{jager2026tabpfn} attains the lowest mean error in BD, FD, IV, and
PX. These results suggest that the quality of the predictive backbone remains consequential.

\end{document}

%% file: math_commands.tex
\usepackage{amsmath,amsfonts,bm}

\def\eqref#1{equation~\ref{#1}}

\def\1{\bm{1}}

\DeclareMathAlphabet{\mathsfit}{\encodingdefault}{\sfdefault}{m}{sl}
\SetMathAlphabet{\mathsfit}{bold}{\encodingdefault}{\sfdefault}{bx}{n}



%% file: sec/1.Introduction.tex
\section{Introduction}
\label{sec:introduction}

\begin{figure*}[ht!]
    \centering
    \setlength{\abovecaptionskip}{2pt}
    \includegraphics[
        trim={0pt 4pt 0pt 0pt},
        clip,
        width=\textwidth
    ]{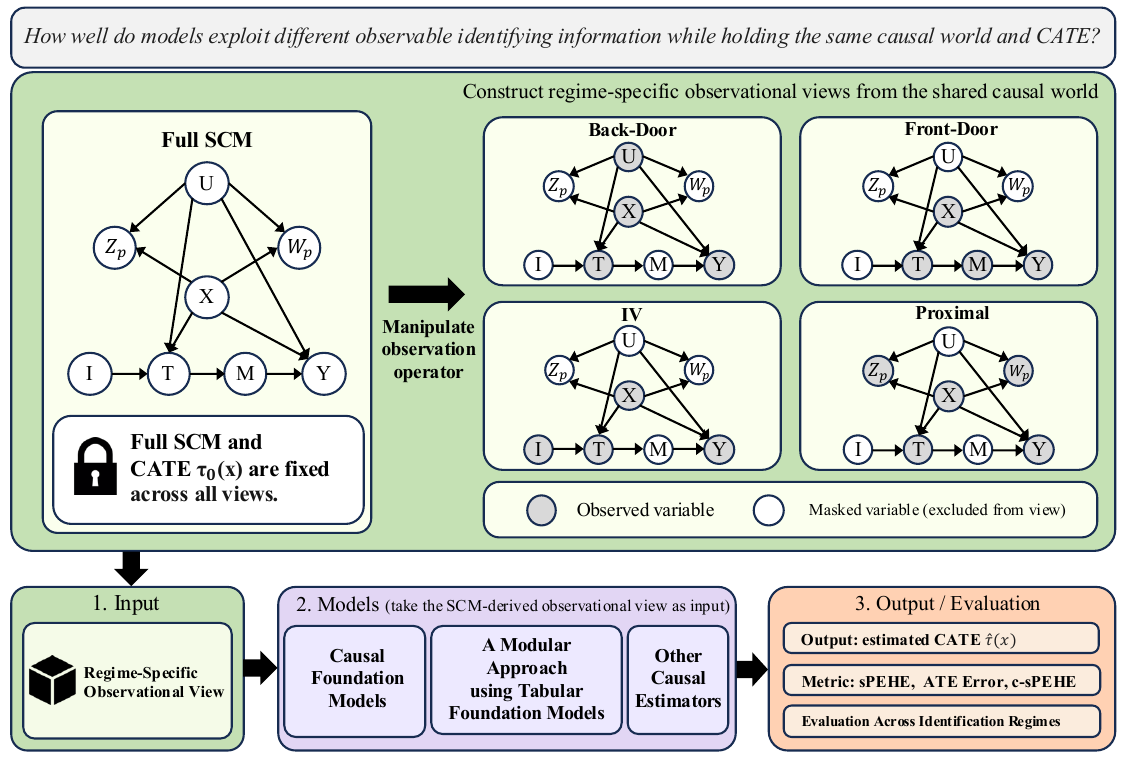}
    \caption{
    \textbf{Overview of \textsc{CausalIDView}.} Each realized SCM is mapped to matched observational views while preserving the query units and target CATE. We can also compare the Hidden Confounding (HC) view, described in Appendix~\ref{app:hidden_confounding}.
    }
    \label{fig:framework}
\end{figure*}

Machine-learning methods for causal effect estimation have become useful estimators by flexibly estimating outcome regressions, propensity scores and other nuisance functions under a specified set of identification assumptions \citep{kunzel2019metalearners,athey2019generalized,shi2019adapting,shalit2017estimating,nie2021quasi,kennedy2023towards}. Recently, Causal Foundation Models (CFMs) have been proposed to amortize causal inference across datasets \citep{balazadeh2026causalpfn,robertson2026pfn,ma2026foundation}. At test time, they take observational datasets as context and
produce causal predictions through in-context learning without
updating pre-trained parameters. This shifts the computational burden from dataset-specific training to shared pre-training, making CFMs attractive for applications requiring causal analyses across multiple datasets. However, current CFMs are trained and evaluated under different observational structures and causal assumptions. It remains unclear how their performance changes when the information available for identification changes.

Existing evaluations do not directly isolate this question.
Across existing CFM studies, pre-training priors and evaluation protocols differ, including the data-generating processes and causal assumptions used for evaluation
\citep{balazadeh2026causalpfn,robertson2026pfn,ma2026foundation}.
\citet{robertson2026pfn} and \citet{ma2026foundation}
consider multiple identification regimes, but their
evaluations use separately constructed SCMs rather than the same underlying system.
This makes it difficult to determine how much performance depends
on the available identifying information.
A controlled comparison should therefore hold the causal world
and target effect fixed while varying only which variables the
estimator observes. A causal world consists of an SCM and one complete data realization, including potential outcomes.

We introduce \textsc{CausalIDView}, a multi-view
benchmark that places causal effect estimators on a common evaluation. For each instance, we hold the SCM realization, query units, and target conditional average treatment effect (CATE) fixed,
while varying only what the estimator observes. As illustrated in Figure~\ref{fig:framework}, each causal world is exposed through matched views supporting back-door adjustment
\citep{pearl1993bayesian}, front-door identification
\citep{pearl2022causal}, instrumental-variable identification
\citep{angrist1996identification}, and proximal identification
\citep{miao2018identifying}, together with a hidden-confounding stress
view. \textsc{CausalIDView} does not assume all views are native to each model since current CFMs were not necessarily pre-trained for every regime. Instead, it characterizes how CFM inference changes under controlled shifts in identifying information and compares CFMs with other causal estimators. We complement CATE accuracy evaluations with semi-synthetic benchmarks and null-effect tests using pre-treatment outcomes from real-world randomized trials.


We further ask whether a strong predictor coupled with explicit
identification can rival CFMs. To this end, we use modular
estimators equipped with the pre-trained tabular foundation model
TabPFN \citep{grinsztajn2026tabpfn, jager2026tabpfn} to estimate the
observable quantities that each regime's identification strategy requires.
No CFM leads in every regime, while TabPFN-based modular estimators
remain competitive with CFMs. Under partial identification, TabPFN-based bound estimators achieve the lowest endpoint errors. Whether structural perturbations
preserve or alter the true CATE, CFMs falter in model-specific ways.
On semi-synthetic benchmarks and real-world null-effect tests,
the best estimator varies by dataset.

Our main contributions are:

\begin{itemize}[
    leftmargin=*,
    labelindent=0pt,
    labelsep=0.5em,
    topsep=2pt,
    itemsep=2pt,
    parsep=0pt,
    partopsep=0pt
]
    \item \textbf{Evaluation ground for causal effect estimators.}
    We introduce a multi-view benchmark that evaluates
    CFMs and causal estimators on various observational views, holding the target fixed.
    
    \item \textbf{Complementary diagnostics.}
    We characterize model-specific failures under structural
    changes, dataset-dependent accuracy, null-effect behavior,
    and evaluation of partial-identification bounds.

    \item \textbf{Effectiveness of modular identification.}
    We demonstrate that combining predictive tabular foundation
    models with explicit causal procedures yields competitive
    point estimates and lower endpoint errors in bound estimation in the
    evaluated settings.
\end{itemize}

%% file: sec/2.Background.tex
\section{Background}
\label{sec:background}

\subsection{CATE and Common Causal Conditions}
\label{sec:target_identification}

\paragraph{CATE.} We use the potential-outcomes framework
\citep{rubin1974estimating}. We consider a binary treatment $T\in\{0,1\}$, an outcome $Y$, and potential outcomes $Y(t)$ under $T=t$. $X$ denotes the observed pre-treatment covariates indexing treatment-effect heterogeneity. The treatment-specific conditional mean and conditional average treatment effect (CATE) are
\begin{equation}
    \mu_t(x)
    :=
    \mathbb{E}\!\left[Y(t)\mid X=x\right],
    \qquad
    \tau_0(x)
    :=
    \mu_1(x)-\mu_0(x)
    =
    \mathbb{E}\!\left[Y(1)-Y(0)\mid X=x\right].
    \label{eq:cate}
\end{equation}


\paragraph{Common causal conditions.}
We maintain consistency together with the standard stable-unit
treatment-value conditions
\citep{rubin1980randomization}.
Consistency requires that the observed outcome equal the
potential outcome under the received treatment, $Y_i=Y_i(T_i)$ for index $i$.
Well-defined treatment versions require each treatment value to
represent a single causally relevant intervention, and
no interference requires that one unit's potential outcomes not
depend on the treatment assignments of other units. The positivity and
support requirements specific to each identification regime are
imposed over the target covariate support and stated in
Appendix~\ref{app:identification_regimes}.

\subsection{Point-Identification Regimes for CATE}
\label{sec:regime_identification}

We focus on four point-identification strategies, depending on which auxiliary causal variables are available. 

{
\setlength{\parskip}{0pt}
\paragraph{Back-Door (BD) identification.}
When the confounder $U$ is observed and $(X,U)$ satisfies conditional exchangeability and positivity, back-door adjustment identifies the target CATE \citep{rosenbaum1983central,pearl2022causal}.
\begin{equation}
\begin{aligned}
    \tau_{\mathrm{BD}}(x)
    &:=
    \int
    \Bigl\{
        \mathbb{E}[Y\mid T=1,X=x,U=u]
        -
        \mathbb{E}[Y\mid T=0,X=x,U=u]
    \Bigr\}
    \,dP(u\mid X=x)
    \nonumber
\end{aligned}
\label{eq:backdoor_functional}
\end{equation}
\paragraph{Front-Door (FD) identification.}
When $U$ is unmeasured but a mediator $M$ satisfying the front-door
criterion is observed, the total treatment effect remains identifiable
\citep{pearl2022causal}. 
For a continuous mediator, the corresponding
conditional front-door functional is

\vspace{-1em}
\begin{equation}
\begin{aligned}
    \tau_{\mathrm{FD}}(x)
    &:=
    \int
    \Bigl\{
        p(m\mid T=1,X=x)
        -
        p(m\mid T=0,X=x)
    \Bigr\} \nonumber \\[-0.25em]
    &\qquad\quad\times
    \left[
        \sum_{a\in\{0,1\}}
        \mathbb{E}[Y\mid M=m,T=a,X=x]\,
        P(T=a\mid X=x)
    \right]dm
    \nonumber
\end{aligned}
\label{eq:frontdoor_functional}
\end{equation}

\paragraph{Instrumental-Variable (IV) identification.}
When a valid binary instrument $I$ is observed, the conditional Wald functional is
\begin{align}
    \tau_{\mathrm{IV}}(x)
    &:=
    \frac{
        \mathbb{E}[Y\mid I=1,X=x]
        -
        \mathbb{E}[Y\mid I=0,X=x]
    }{
        \mathbb{E}[T\mid I=1,X=x]
        -
        \mathbb{E}[T\mid I=0,X=x]
    }
    \nonumber
    \label{eq:iv_functional}
\end{align}
The conditional Wald ratio generally identifies a conditional local average treatment effect under the standard relevance, exogeneity, exclusion, and monotonicity conditions~\citep{angrist1995identification,angrist1996identification}.

\paragraph{Proximal Causal (PX) identification.}
When $U$ is unmeasured but the confounding proxies
$(Z_{\mathrm{p}},W_{\mathrm{p}})$ are observed, let $h_t(w,x)$ be an
outcome confounding bridge satisfying
  $ \mathbb{E}[Y\mid Z_{\mathrm{p}}=z,T=t,X=x]
    =
    \mathbb{E}
    \!\left[
        h_t(W_{\mathrm{p}},x)
        \mid
        Z_{\mathrm{p}}=z,T=t,X=x
    \right].
    \label{eq:proximal_bridge}$
Under the corresponding proxy-independence, bridge-existence, and
completeness conditions, the proximal g-formula \citep{miao2018identifying,tchetgen2024introduction,
sverdrup2023proximal} identifies
\begin{align}
    \tau_{\mathrm{PX}}(x)
    :=
    \mathbb{E}
    \!\left[
        h_1(W_{\mathrm{p}},x)
        -
        h_0(W_{\mathrm{p}},x)
        \mid X=x
    \right]
    \nonumber
\end{align}


Under the benchmark's maintained regime-specific conditions,
with the shared-target construction described in
Section~\ref{sec:paired_worlds},
all four identification strategies recover the same target CATE:
\begin{equation}
    \boxed{
        \tau_{\mathrm{BD}}(x)
        =
        \tau_{\mathrm{FD}}(x)
        =
        \tau_{\mathrm{IV}}(x)
        =
        \tau_{\mathrm{PX}}(x)
        =
        \tau_0(x)
    }
    \label{eq:shared_identified_target}
\end{equation}
Appendix~\ref{app:identification_regimes} formally provides the assumptions and derivations
for each identification regime.

%% file: sec/3.Related_Work.tex
\section{Related Work}
{
\setlength{\parskip}{0pt}
\paragraph{Identification of Causal Effects} 

Causal identification concerns whether a target causal estimand is uniquely determined by the observed-data distribution under maintained causal assumptions. In graphical causal models, assumptions encoded by the causal structure can render interventional or counterfactual quantities identifiable from observational data. In general, complete identification procedures characterize when causal queries can be uniquely recovered from lower levels of the causal hierarchy and provide graphical certificates when such recovery is impossible \citep{JMLR:v9:shpitser08a,shpitser2020identification}. When multiple causal models remain observationally indistinguishable yet imply different values of the target estimand, point identification fails. Rather than treating such cases as entirely uninformative, partial identification characterizes the set of causal effects compatible with the observed distribution and maintained assumptions, often through lower and upper bounds. This perspective has a long history in econometrics and causal inference, including bounds for treatment effects under imperfect compliance \citep{balke1997bounds} and the broader theory of partially identified probability distributions \citep{manski2003partial}.

\paragraph{Causal Foundation Models for Treatment-Effect Estimation}

Recent work explores amortized causal effect estimation through pre-trained foundation models. Do-PFN~\citep{robertson2026pfn} pre-trains a Prior-Data Fitted Network (PFN)~\citep{muller2021transformers} over diverse causal structures using paired observational and interventional samples to predict conditional interventional distributions. CausalPFN~\citep{balazadeh2026causalpfn} focuses on treatment-effect estimation under strong ignorability across heterogeneous data-generating processes, and CausalFM~\citep{ma2026foundation} explicitly separates identification from estimation through identification-aware SCM priors covering back-door, front-door, and instrumental-variable settings. Causal foundation models have largely targeted point estimates or prior-dependent posterior predictions, with recent work extending to partial identification~\citep{bellot2026foundation}.

\paragraph{Benchmarks for Causal Effect Estimation}

Existing benchmarks for causal effect estimation rely largely on synthetic or semi-synthetic data and evaluate estimators under controlled variations in data-generating processes. Widely used benchmarks such as IHDP~\citep{Hill01012011} and the ACIC challenges~\citep{dorie2019automated, hahn2019atlantic} vary outcome functions, treatment assignment, overlap, and effect heterogeneity, while RealCause~\citep{neal2020realcause} improves realism by learning generative models from observational datasets. However, these benchmarks largely vary the DGP within a fixed identification regime rather than evaluating estimators across different identification regimes. This motivates benchmarks spanning diverse SCMs and identification regimes, analogous to broad generalization benchmarks such as TabArena~\citep{erickson2026tabarena}.
}

%% file: sec/4.Data_Curation.tex
\section{Data Curation}
\label{sec:data_curation}

\textsc{CausalIDView} is designed to contrast the performance of estimators targeting the same causal effect as the available identifying information changes.
Rather than generating a separate dataset for each identification
regime, we generate each complete causal world once and derive
multiple observational views from it. Within each world, the SCM
realization, query units, and target CATE remain fixed, while only the variables available to the estimator change.

To complement the analysis on our controlled synthetic benchmark, we examine baseline performance on semi-synthetic and real-world datasets that reflect characteristics of empirical data. Using semi-synthetic datasets with more realistic covariate distributions, treatment assignments and outcome variability, we evaluate CATE estimation accuracy against the available ground-truth effects. In the real-world setting, where ground-truth CATE is not  observed but is theoretically known to be zero, we investigate whether models introduce  treatment-effect bias into their estimates.

\subsection{Synthetic Dataset}

\subsubsection{Generating Shared Causal Worlds}
\label{sec:paired_worlds}

\paragraph{Shared causal structure.}
Each world is generated by the causal graph in
Figure~\ref{fig:framework}.
The observational views are projections of a shared generating
process, not independently constructed regime-specific datasets.
The confounder $U$ affects treatment assignment and outcome
levels, while the treatment effect on $Y$ is mediated entirely
through $M$. A randomized binary instrument $I$ affects the outcome only
through treatment, and the pre-treatment proxies $(Z_p,W_p)$
provide noisy measurements of $U$ without directly affecting
treatment or outcome. All variables are generated jointly before any observational view
is constructed.

\paragraph{Shared target CATE.}
To realize the common-target identity in ~\eqref{eq:shared_identified_target}, we restrict $U$ to enter both potential outcomes through the same additive term
$\gamma_U(X)U$, rather than modifying the treatment gain.
The mediator has a fixed treatment-induced shift $\delta$, and
its contribution to the outcome has coefficient $\beta(X)$.
We use shared unit-level mediator and outcome noises across
treatment arms.
Within a fixed world, the resulting unit-level effect and its
conditional expectation satisfy
\begin{equation}
\begin{aligned}
    Y_i(1)-Y_i(0)
    &=
    \beta(X_i)\{M_i(1)-M_i(0)\}
    =
    \delta\beta(X_i), \\
    \tau_0(x)
    &=
    \mathbb{E}[Y_i(1)-Y_i(0)\mid X_i=x]
    =
    \delta\beta(x).
\end{aligned}
\label{eq:curation_shared_effect}
\end{equation}
Under this construction, the unit-level treatment effect is
determined entirely by $X$:
\[
Y_i(1)-Y_i(0)=\tau_0(X_i).
\]
Therefore, the conditional average treatment effect in ~\eqref{eq:cate} is exactly the calibrated effect surface
$\tau_0(x)$, and the same target is attached to every observational
view of a given causal world. The restriction also removes variation in treatment gains across
latent states or compliance types at fixed $X$, so the conditional
IV complier effect coincides with this target CATE.

We then construct the remaining mechanisms of the SCM so that each
observational view satisfies the assumptions required to identify
this same target. Although the observed variables and identifying functionals
differ across views, each point-identified view recovers the same
$\tau_0(x)$.
Appendix~\ref{app:identification_regimes} provides the assumptions and derivations and Appendix~\ref{sec:data-validation} reports construction and identification audits.


\paragraph{Confounding in the observational contrast.}
The gain restriction does not eliminate confounding from the
observed treatment-group contrast.
Under this SCM,
\begin{equation}
\begin{aligned}
    &\mathbb{E}[Y \mid T=1,X=x]
    -\mathbb{E}[Y \mid T=0,X=x] \\
    &\quad =
    \tau_0(x)
    +
    \underbrace{
        \gamma_U(x)
        \bigl\{
            \mathbb{E}[U \mid T=1,X=x]
            -
            \mathbb{E}[U \mid T=0,X=x]
        \bigr\}
    }_{\text{confounding bias}}.
\end{aligned}
\label{eq:curation_confounding_bias}
\end{equation}
The common confounding term cancels between the two potential
outcomes of the same unit, but not between treatment groups,
because treatment selection changes the conditional
distribution of $U$. 
The construction aligns the causal targets across regimes
without eliminating confounding in either the level or shape
of the observational contrast.

\subsubsection{Observational Views}
\label{sec:view_curation}

After generating a complete world, we derive multiple
observational views by changing only which auxiliary variables
are available to the estimator.
The BD, FD, IV, and PX views retain $U$, $M$, $I$, and
$(Z_p,W_p)$, respectively, alongside $(X,T,Y)$.

We assign a single disjoint context--query split, denoted by
$\mathcal{C}$ and $\mathcal{Q}$, and reuse it across all views.
For view $r$, let $V^{(r)}$ denote its retained auxiliary variables.
The observational context is
\begin{equation}
    \mathcal{D}_{\mathrm{obs}}^{(r)}
    =
    \left\{
        (X_i,T_i,Y_i,V_i^{(r)})
        :
        i\in\mathcal{C}
    \right\}.
    \label{eq:view_dataset}
\end{equation}

Excluding a variable from a view changes its availability to the
estimator, not its role in generating the data.
In particular, we do not resample treatment assignments,
outcomes, structural parameters, or unit-level noise when
constructing a new view.
All views within a world are evaluated on the same query targets
$\{\tau_0(X_i):i\in\mathcal{Q}\}$.
Auxiliary variables supply information for estimation without
redefining the target conditioning set.

Consequently, within-world performance differences across views
are not driven by different realized outcomes, query populations,
or oracle effect values. This pairing compares the performance of models on the same
realized causal problem while varying the information available
for identification. Potential outcomes and oracle effects are reserved for evaluation
and data validation.

\subsection{Semi-Synthetic Dataset}
We evaluate the CATE estimation performance of existing back-door baselines using ACIC 2016~\citep{dorie2019automated} and LaLonde-PSID/CPS datasets~\citep{lalonde1986evaluating, dehejia1999causal, dehejia2002propensity} based on RealCause~\citep{neal2020realcause}. ACIC 2016 is a benchmark that combines real covariates from the Collaborative Perinatal Project with researcher-designed treatment assignment and outcome-generating mechanisms. By preserving realistic covariate distributions, it enables the evaluation of estimation performance under diverse data-generating conditions, including nonlinearity, varying degrees of overlap, and treatment-effect heterogeneity. The RealCause-based LaLonde-PSID/CPS generates data from real covariates together with conditional treatment-assignment and outcome distributions learned from observational data. The original data concern the effect of participation in a job-training program on subsequent earnings and consist of the treatment group from the National Supported Work program paired with nonexperimental comparison groups drawn from PSID and CPS, respectively. RealCause fits generative models to the observational data under an assumed causal structure, providing ground-truth CATEs defined by the resulting data-generating process. We use back-door baselines as the relevant comparison methods since both benchmarks assume that confounding between treatment and outcome can be sufficiently controlled using the observed covariates. Detailed experimental settings are provided in Appendix~\ref{app:semi-synthetic}.

\subsection{Real-World Dataset}
Estimating causal effects in real-world settings is an important problem, yet the ground-truth counterfactuals needed to evaluate treatment-effect estimates are generally unavailable. To address this limitation, we construct a real-world evaluation setting with known ground-truth CATEs by using variables measured before treatment assignment in randomized controlled trials (RCTs) as negative control outcomes. A negative control outcome is an outcome that cannot be affected by the treatment of interest~\citep{arnold2016negative}. In particular, an outcome realized before treatment assignment cannot be altered by a subsequent treatment.~\citet{ashby2025negative} formalize this property in the potential-outcome framework as a zero treatment effect for every individual. Thus, for a pre-treatment outcome $Y^{\mathrm{pre}}$, we have $Y_i^{\mathrm{pre}}(1)=Y_i^{\mathrm{pre}}(0)$, implying $\tau_0(x)=\mathbb{E}[Y^{\mathrm{pre}}(1)-Y^{\mathrm{pre}}(0)\mid X=x]=0$ for all $x$.

Following this principle, we use three real-world RCT datasets. For ACTG 175~\citep{hammer1996trial}, we focus on the treatment arms randomized to zidovudine (AZT) monotherapy and AZT+didanosine (ddI) combination therapy. Baseline CD4 count, measured before treatment assignment, serves as the outcome. The Pennsylvania Reemployment Bonus Demonstration~\citep{corson1992pennsylvania} compares the control group with a reemployment-bonus treatment arm. We use earnings measured before randomized bonus assignment as the outcome. In the Illinois Unemployment Insurance Incentive Experiment~\citep{woodbury1987bonuses}, the treatment contrast is between the control group and claimants offered a reemployment bonus. Base-period earnings measured before treatment assignment serve as the outcome. Appendix~\ref{app:real-world} provides the conditions and proofs under which the null effect remains valid when treatment arms are selected in multi-treatment RCTs.

%% file: sec/5.Experimental_Design.tex
\section{Experimental Design}
\label{sec:experiments}

\subsection{Metrics}
\label{sec:metrics}

For query units $\{x_i\}_{i=1}^{N_q}$, we evaluate CATE estimation using
\[
\mathrm{sPEHE}
=
\sqrt{
    \frac{1}{N_q}
    \sum_{i=1}^{N_q}
    \bigl(\widehat{\tau}(x_i)-\tau_0(x_i)\bigr)^2
}.
\]
Let $\tau_{\mathrm{ATE},\mathcal Q}=N_q^{-1}\sum_{i=1}^{N_q}\tau_0(x_i)$ and $\widehat{\tau}_{\mathrm{ATE},\mathcal Q}=N_q^{-1}\sum_{i=1}^{N_q}\widehat{\tau}(x_i)$. We additionally report the ATE error $|\widehat{\tau}_{\mathrm{ATE},\mathcal Q}-\tau_{\mathrm{ATE},\mathcal Q}|$ and centered sPEHE (c-sPEHE), which separate errors in the query-mean effect from errors in the centered CATE function with $\mathrm{sPEHE}^2=(\widehat{\tau}_{\mathrm{ATE},\mathcal Q}-\tau_{\mathrm{ATE},\mathcal Q})^2+\mathrm{c\text{-}sPEHE}^2$. Appendix~\ref{app:evaluation_metrics} provides the detailed definitions and derivation.

Standard point-estimation metrics do not measure whether an estimator
responds appropriately to structural changes. We therefore consider
Stay and Move conditions. Stay changes the structural mechanism while
preserving the target CATE,
$\tau_0^{\mathrm{stay}}(x_i)=\tau_0(x_i)$, so
$E_{\mathrm{stay}}$ measures spurious changes in the estimated CATE
under a CATE-preserving intervention. Move instead changes the target
CATE itself and $E_{\mathrm{move}}$ measures how accurately the
corresponding change in the estimated CATE tracks the true CATE change.
For a function $f$, let
$\|f\|_{\mathcal Q}
:=
\sqrt{N_q^{-1}\sum_{i=1}^{N_q}f(x_i)^2}$.
Both metrics are dimensionless and lower values are better.
\[
E_{\mathrm{stay}}
=
\frac{
    \|\widehat{\tau}^{\mathrm{stay}}-\widehat{\tau}\|_{\mathcal Q}
}{
    \|\tau_0\|_{\mathcal Q}
},
\qquad
E_{\mathrm{move}}
=
\frac{
    \|(\widehat{\tau}^{\mathrm{move}}-\widehat{\tau})
    -(\tau_0^{\mathrm{move}}-\tau_0)\|_{\mathcal Q}
}{
    \|\tau_0^{\mathrm{move}}-\tau_0\|_{\mathcal Q}
}.
\]

\subsection{Baselines}
\paragraph{Causal foundation models.}
We evaluate CausalPFN, Do-PFN, and CausalFM
\citep{balazadeh2026causalpfn,robertson2026pfn,ma2026foundation}.
In these configurations, CausalPFN and Do-PFN use auxiliary variables as ordinary features, without IV/proximal role annotations or specialized wrappers. The front-door \emph{X-only} variants omit $M$ from both context and query, avoiding conditioning total-effect predictions on the factual mediator.

\paragraph{Modular approaches.}
Our modular approaches pair supervised machine-learning models with
explicit causal estimation procedures for the corresponding
identification regimes. The predictive models estimate the outcome, propensity, or other
nuisance quantities required by each procedure, while the causal
estimation step determines how predictions are combined into
treatment-effect estimates.

In the back-door setting, we use S- and X-Learners
\citep{kunzel2019metalearners} and a DR-Learner
\citep{kennedy2023towards}.
Unlike the S- and X-Learners, the DR-Learner constructs
doubly robust pseudo-outcomes by augmenting outcome contrasts with
inverse-propensity-weighted residual corrections.

Under front-door identification, we replace the predictive components
of the plug-in estimator, including the treatment propensity, mediator,
and outcome models, while retaining the front-door averaging procedure.
The IV variant estimates the outcome mean and treatment probability
under each instrument value and combines their contrasts through the
conditional Wald ratio.
For the proximal setting, we use a P-Learner
\citep{sverdrup2023proximal}, replacing the predictive models used for
bridge estimation and final pseudo-outcome regression.

We instantiate these procedures with predictive models including
TabPFN-v3.5 \citep{jager2026tabpfn}, XGBoost
\citep{chen2016xgboost}, neural networks, and ExtraTrees \citep{geurts2006extremely} as applicable.

\paragraph{Other estimators.}
We additionally include regime-specific causal estimators that do not
belong to either the CFM or modular predictor--procedure categories,
including ForestDRIV and KIV for instrumental-variable estimation
\citep{syrgkanis2019machine,singh2019kernel}.

Implementation details of all baselines are provided in Appendix~\ref{sec:baselines}.

%% file: sec/6.Result.tex
\section{Results}
\label{sec:result}
\begin{figure*}[t]
    \centering
    \setlength{\abovecaptionskip}{2pt}
    \setlength{\parskip}{0pt}
    \includegraphics[width=\textwidth]{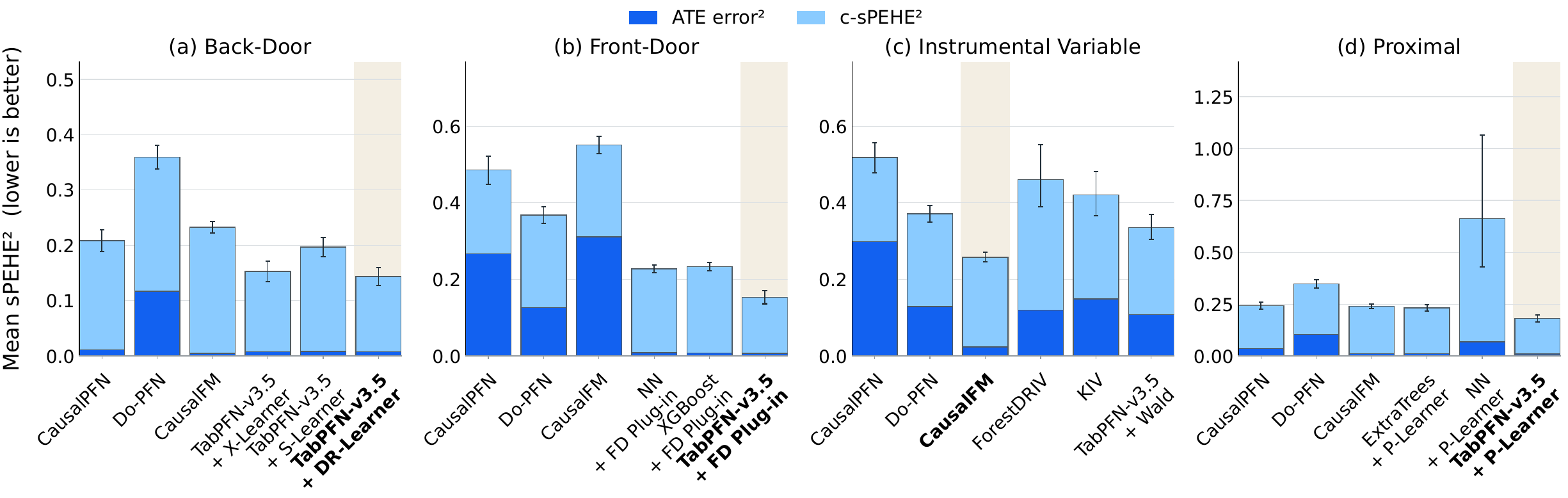}
    \caption{
    \textbf{Point estimation across observational views.}
    We compare the evaluated model configurations under
    (a) Back-Door, (b) Front-Door, (c) IV, and
    (d) Proximal views.
    Bars show mean $\mathrm{sPEHE}^{2}$ across 40 SCM worlds,
    decomposed into mean squared query-sample ATE error
    (dark blue) and mean $\mathrm{c\text{-}sPEHE}^{2}$
    (light blue).
    Shading marks the lowest observed mean in each panel.
    CausalPFN and Do-PFN receive the auxiliary variables available in each view as ordinary covariates, except for the post-treatment mediator.
    }
    \label{fig:regime_point_estimation}
\end{figure*}

\paragraph{RQ1: How does performance vary across observational views?}
\label{sec:rq_views}
Figure~\ref{fig:regime_point_estimation} compares the evaluated
model--input configurations while holding the causal world and target
CATE fixed, and shows that the lowest-error configuration differs
across observational views.
The TabPFN-based DR-Learner and front-door plug-in attain the lowest
mean $\mathrm{sPEHE}^2$ in back-door and front-door, respectively,
whereas CausalFM leads in IV.
In the proximal view, the TabPFN-based P-Learner attains the lowest
mean $\mathrm{sPEHE}^2$ among the evaluated configurations.
The decomposition of $\mathrm{sPEHE}^2$ distinguishes errors in the
query-mean effect from errors in the centered CATE function and shows
that their contributions vary across model--view configurations.
Centered CATE error dominates most back-door results and several
proximal results, whereas mean-effect error is substantial for
particular front-door and IV configurations.
Thus, a single total-error score can obscure distinct estimation failures.
The back-door and FD comparisons illustrate the practical value of
combining a predictive model with an explicit causal procedure,
while the IV comparison shows that this advantage does not hold
uniformly across regimes. 

\begin{figure*}[t]
    \centering
    \setlength{\abovecaptionskip}{2pt}
    \setlength{\parskip}{0pt}
    \includegraphics[width=\textwidth]{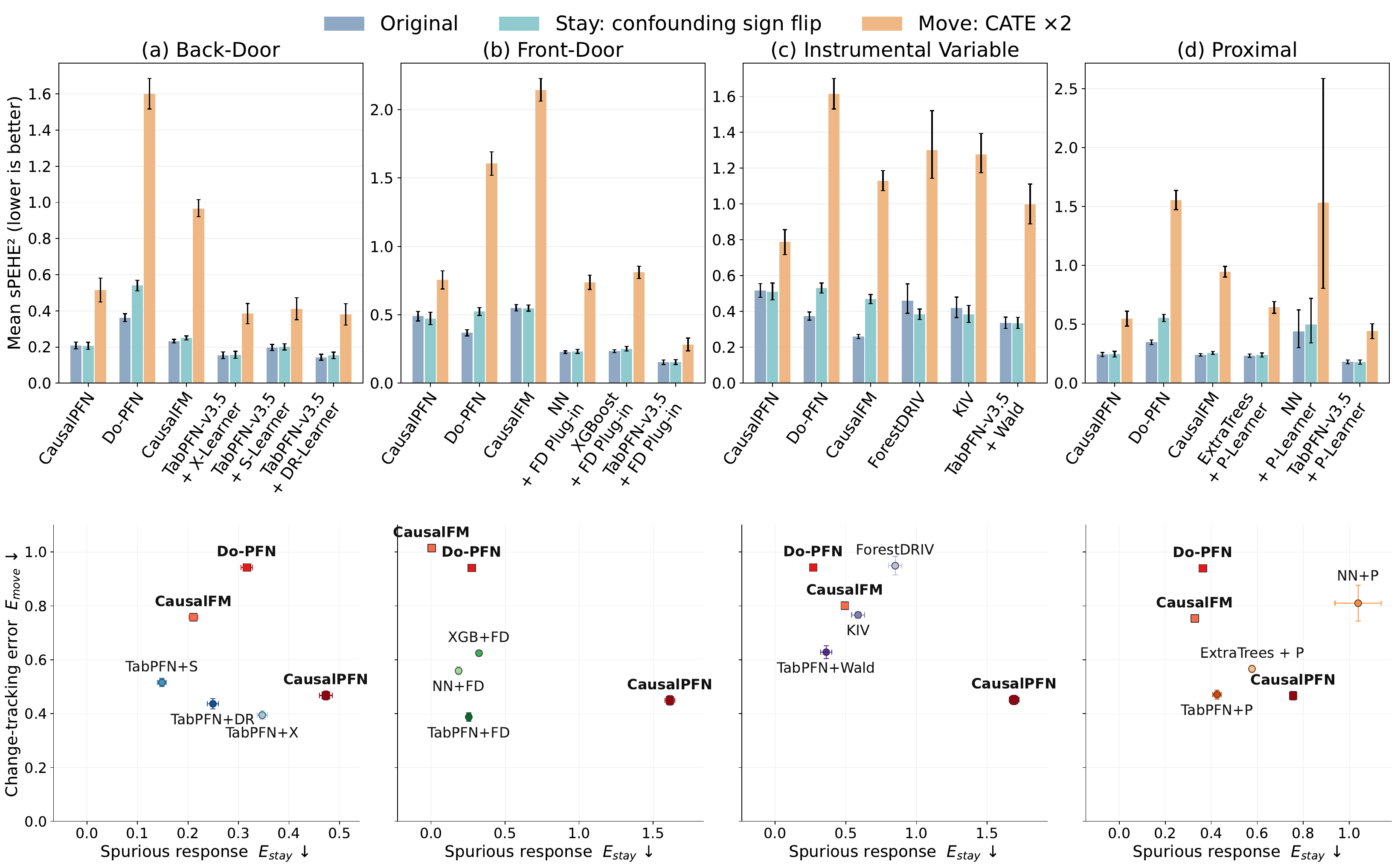}
    \caption{
    \textbf{Structural-response audit across observational views.}
    We evaluate model responses under
    (a) Back-Door, (b) Front-Door, (c) IV, and (d) Proximal views.
    \textbf{Top:}
    Bars show mean $\mathrm{sPEHE}^{2}$ under the Original, Stay, and Move conditions.
    \textbf{Bottom:}
    Each point represents an evaluated model configuration,
    positioned by its mean spurious-response error
    $E_{\mathrm{stay}}$ on the horizontal axis and mean
    change-tracking error $E_{\mathrm{move}}$ on the vertical axis.
    }
    \label{fig:structural_response}
\end{figure*}

\paragraph{RQ2: Do estimators respond appropriately to structural changes?}
\label{sec:rq_response}
Figure~\ref{fig:structural_response} examines whether estimators
respond to changes in the target causal effect rather than to
incidental structural changes.
Starting from 
$Y(t)=\mu(X)+\gamma_U(X)U+\beta(X)M(t)+\epsilon_Y$,
we construct a Stay intervention by reversing the confounding
contribution, yielding $\tau_0^{\mathrm{stay}}(x)=\tau_0(x)$,
and a Move intervention by doubling the mediator--outcome response,
yielding $\tau_0^{\mathrm{move}}(x)=2\tau_0(x)$.
An appropriate response therefore requires invariance under Stay
and accurate effect-change tracking under Move.
The resulting response map reveals qualitatively different CFM failures.
CausalPFN shows relatively large Stay errors, indicating that its
predictions can change even when the target CATE is preserved.
On the other hand, Do-PFN shows consistently large Move errors
with poor tracking when the CATE itself changes.
CausalFM exhibits a view-dependent mixture of these behaviors.
In contrast, the TabPFN-v3.5 modular estimators generally occupy
a more balanced region with lower errors on both axes, suggesting
that combining a strong predictor with an explicit identification
rule better preserves invariance while tracking genuine effect changes.
These results motivate evaluating structural responses alongside
point-estimation accuracy and developing CFMs that more explicitly
distinguish estimand-preserving structural shifts from those that
alter the causal effect.

\begin{table*}[t]
\centering
\caption{
\textbf{Back-door evaluation on semi-synthetic benchmarks.}
Mean squared errors over 40 repeated realizations.
Lower is better.
For each dataset--metric column, the lowest value is shown in
\textcolor{red}{red} and the second-lowest value in
\textcolor{orange}{orange}.
Values are reported as mean $\pm$ standard deviation.
LaLonde-PSID and LaLonde-CPS results are shown in units of $\times 10^{6}$.
}
\label{tab:semisynthetic_backdoor}

\setlength{\tabcolsep}{3.5pt}
\renewcommand{\arraystretch}{1.08}
\small

\resizebox{\textwidth}{!}{
\begin{tabular}{lccc|ccc|ccc}
\toprule
\multirow{2}{*}{\textbf{Estimator}}
&
\multicolumn{3}{c|}{\textbf{ACIC 2016}}
&
\multicolumn{3}{c|}{\textbf{LaLonde-PSID} ($\times 10^6$)}
&
\multicolumn{3}{c}{\textbf{LaLonde-CPS} ($\times 10^6$)}
\\

\cmidrule(lr){2-4}
\cmidrule(lr){5-7}
\cmidrule(lr){8-10}

&
$\mathrm{sPEHE}^2$
&
$\mathrm{ATE\ error}^2$
&
$\mathrm{c\text{-}sPEHE}^2$
&
$\mathrm{sPEHE}^2$
&
$\mathrm{ATE\ error}^2$
&
$\mathrm{c\text{-}sPEHE}^2$
&
$\mathrm{sPEHE}^2$
&
$\mathrm{ATE\ error}^2$
&
$\mathrm{c\text{-}sPEHE}^2$
\\
\midrule

CausalPFN
& $3.24$ {\scriptsize $\pm 3.45$}
& $0.1074$ {\scriptsize $\pm 0.2224$}
& $3.13$ {\scriptsize $\pm 3.34$}
& $44.0$ {\scriptsize $\pm 61.5$}
& $27.2$ {\scriptsize $\pm 41.7$}
& \textcolor{orange}{$16.7$ {\scriptsize $\pm 21.6$}}
& \textcolor{red}{$11.2$ {\scriptsize $\pm 8.03$}}
& \textcolor{red}{$7.55$ {\scriptsize $\pm 8.10$}}
& \textcolor{red}{$3.65$ {\scriptsize $\pm 0.782$}}
\\

Do-PFN
& $31.48$ {\scriptsize $\pm 20.02$}
& $12.19$ {\scriptsize $\pm 9.29$}
& $19.29$ {\scriptsize $\pm 18.65$}
& $373$ {\scriptsize $\pm 37.2$}
& $213$ {\scriptsize $\pm 34.8$}
& $160$ {\scriptsize $\pm 7.20$}
& $80.6$ {\scriptsize $\pm 4.70$}
& $43.6$ {\scriptsize $\pm 3.81$}
& $37.0$ {\scriptsize $\pm 1.77$}
\\

CausalFM
& $23.56$ {\scriptsize $\pm 17.52$}
& $5.49$ {\scriptsize $\pm 4.84$}
& $18.08$ {\scriptsize $\pm 17.49$}
& $423$ {\scriptsize $\pm 7.50$}
& $212$ {\scriptsize $\pm 6.85$}
& $211$ {\scriptsize $\pm 4.49$}
& $85.6$ {\scriptsize $\pm 1.31$}
& $44.4$ {\scriptsize $\pm 0.711$}
& $41.2$ {\scriptsize $\pm 1.42$}
\\

Causal Tree
& $11.43$ {\scriptsize $\pm 9.75$}
& $1.30$ {\scriptsize $\pm 1.41$}
& $10.14$ {\scriptsize $\pm 9.86$}
& $181$ {\scriptsize $\pm 23.7$}
& \textcolor{red}{$2.23$ {\scriptsize $\pm 4.34$}}
& $179$ {\scriptsize $\pm 22.9$}
& $61.2$ {\scriptsize $\pm 8.83$}
& $12.4$ {\scriptsize $\pm 8.83$}
& $48.8$ {\scriptsize $\pm 0.0000$}
\\

TARNet
& $9.44$ {\scriptsize $\pm 8.15$}
& $0.2352$ {\scriptsize $\pm 0.3589$}
& $9.20$ {\scriptsize $\pm 8.09$}
& $131$ {\scriptsize $\pm 212$}
& $46.6$ {\scriptsize $\pm 90.2$}
& $84.4$ {\scriptsize $\pm 130$}
& $49.2$ {\scriptsize $\pm 21.3$}
& $23.4$ {\scriptsize $\pm 6.55$}
& $25.7$ {\scriptsize $\pm 21.2$}
\\

GRF / Causal Forest
& $8.30$ {\scriptsize $\pm 9.66$}
& $0.1903$ {\scriptsize $\pm 0.2633$}
& $8.11$ {\scriptsize $\pm 9.68$}
& $439$ {\scriptsize $\pm 62.8$}
& $247$ {\scriptsize $\pm 58.6$}
& $192$ {\scriptsize $\pm 10.5$}
& $96.3$ {\scriptsize $\pm 19.9$}
& $47.7$ {\scriptsize $\pm 19.6$}
& $48.6$ {\scriptsize $\pm 1.56$}
\\

CFRNet
& $11.43$ {\scriptsize $\pm 9.38$}
& $0.5777$ {\scriptsize $\pm 0.7625$}
& $10.85$ {\scriptsize $\pm 9.16$}
& \textcolor{red}{$18.4$ {\scriptsize $\pm 33.6$}}
& \textcolor{orange}{$3.77$ {\scriptsize $\pm 6.94$}}
& \textcolor{red}{$14.7$ {\scriptsize $\pm 28.3$}}
& $18.7$ {\scriptsize $\pm 11.5$}
& $11.0$ {\scriptsize $\pm 8.39$}
& $7.64$ {\scriptsize $\pm 8.62$}
\\

TabPFN-v3.5 + X-Learner
& \textcolor{red}{$1.11$ {\scriptsize $\pm 1.56$}}
& \textcolor{orange}{$0.0258$ {\scriptsize $\pm 0.0417$}}
& \textcolor{red}{$1.09$ {\scriptsize $\pm 1.55$}}
& $456$ {\scriptsize $\pm 295$}
& $223$ {\scriptsize $\pm 180$}
& $233$ {\scriptsize $\pm 127$}
& $145$ {\scriptsize $\pm 132$}
& $86.7$ {\scriptsize $\pm 107$}
& $58.2$ {\scriptsize $\pm 26.7$}
\\

TabPFN-v3.5 + S-Learner
& $1.25$ {\scriptsize $\pm 2.03$}
& $0.0282$ {\scriptsize $\pm 0.0290$}
& $1.23$ {\scriptsize $\pm 2.03$}
& $415$ {\scriptsize $\pm 116$}
& $200$ {\scriptsize $\pm 80.2$}
& $215$ {\scriptsize $\pm 44.3$}
& $74.6$ {\scriptsize $\pm 9.64$}
& $37.0$ {\scriptsize $\pm 8.13$}
& $37.6$ {\scriptsize $\pm 4.74$}
\\

TabPFN-v3.5 + DR-Learner
& \textcolor{orange}{$1.17$ {\scriptsize $\pm 1.54$}}
& \textcolor{red}{$0.0222$ {\scriptsize $\pm 0.0280$}}
& \textcolor{orange}{$1.15$ {\scriptsize $\pm 1.54$}}
& \textcolor{orange}{$42.8$ {\scriptsize $\pm 63.8$}}
& $20.0$ {\scriptsize $\pm 33.0$}
& $22.8$ {\scriptsize $\pm 32.6$}
& \textcolor{orange}{$15.3$ {\scriptsize $\pm 23.5$}}
& \textcolor{orange}{$9.10$ {\scriptsize $\pm 19.3$}}
& \textcolor{orange}{$6.23$ {\scriptsize $\pm 5.46$}}
\\

\bottomrule
\end{tabular}
}
\end{table*}

\begin{table*}[t]
\centering
\caption{
\textbf{RCT negative-control null-CATE benchmarks.}
Mean squared null-CATE error over 40 repeated context/query realizations.
Lower is better.
For each dataset--metric column, the lowest value is shown in
\textcolor{bestred}{red} and the second-lowest distinct value in
\textcolor{secondorange}{orange}. Tied values receive the same color.
Each dataset uses its primary context size $n_C$.}
\label{tab:null_cate_rct}

\resizebox{\textwidth}{!}{
\begin{tabular}{lccc|ccc|ccc}
\toprule
\multirow{2}{*}{\textbf{Estimator}}
&
\multicolumn{3}{c|}{\textbf{ACTG 175} ($n_C=768$)}
&
\multicolumn{3}{c|}{\textbf{Pennsylvania} ($n_C=1024$)}
&
\multicolumn{3}{c}{\textbf{Illinois} ($n_C=2048$)}
\\

\cmidrule(lr){2-4}
\cmidrule(lr){5-7}
\cmidrule(lr){8-10}

&
$\mathrm{sPEHE}^2$
&
$\mathrm{ATE\ error}^2$
&
$\mathrm{c\text{-}sPEHE}^2$
&
$\mathrm{sPEHE}^2$
&
$\mathrm{ATE\ error}^2$
&
$\mathrm{c\text{-}sPEHE}^2$
&
$\mathrm{sPEHE}^2$
&
$\mathrm{ATE\ error}^2$
&
$\mathrm{c\text{-}sPEHE}^2$
\\
\midrule

CausalPFN
& $0.0146\,{\scriptstyle \pm 0.0086}$
& $0.0028\,{\scriptstyle \pm 0.0035}$
& $0.0118\,{\scriptstyle \pm 0.0068}$
& $0.0166\,{\scriptstyle \pm 0.0089}$
& $0.0030\,{\scriptstyle \pm 0.0036}$
& $0.0136\,{\scriptstyle \pm 0.0064}$
& \textcolor{secondorange}{$0.0051\,{\scriptstyle \pm 0.0040}$}
& $0.0011\,{\scriptstyle \pm 0.0019}$
& \textcolor{secondorange}{$0.0039\,{\scriptstyle \pm 0.0031}$}
\\

Do-PFN
& $0.0203\,{\scriptstyle \pm 0.0113}$
& $0.0113\,{\scriptstyle \pm 0.0094}$
& $0.0090\,{\scriptstyle \pm 0.0062}$
& $0.0177\,{\scriptstyle \pm 0.0080}$
& $0.0099\,{\scriptstyle \pm 0.0052}$
& $0.0077\,{\scriptstyle \pm 0.0037}$
& $0.1249\,{\scriptstyle \pm 0.0220}$
& $0.0960\,{\scriptstyle \pm 0.0210}$
& $0.0289\,{\scriptstyle \pm 0.0069}$
\\

CausalFM
& \textcolor{secondorange}{$0.0053\,{\scriptstyle \pm 0.0028}$}
& \textcolor{secondorange}{$0.0009\,{\scriptstyle \pm 0.0010}$}
& \textcolor{secondorange}{$0.0045\,{\scriptstyle \pm 0.0029}$}
& \textcolor{secondorange}{$0.0023\,{\scriptstyle \pm 0.0011}$}
& \textcolor{secondorange}{$0.0016\,{\scriptstyle \pm 0.0010}$}
& \textcolor{bestred}{$0.0007\,{\scriptstyle \pm 0.0003}$}
& \textcolor{bestred}{$0.0016\,{\scriptstyle \pm 0.0008}$}
& \textcolor{bestred}{$0.0004\,{\scriptstyle \pm 0.0005}$}
& \textcolor{bestred}{$0.0011\,{\scriptstyle \pm 0.0005}$}
\\

Causal Tree
& $0.0416\,{\scriptstyle \pm 0.0485}$
& $0.0022\,{\scriptstyle \pm 0.0028}$
& $0.0394\,{\scriptstyle \pm 0.0484}$
& $0.0173\,{\scriptstyle \pm 0.0352}$
& $0.0042\,{\scriptstyle \pm 0.0058}$
& $0.0131\,{\scriptstyle \pm 0.0308}$
& $0.0082\,{\scriptstyle \pm 0.0197}$
& $0.0017\,{\scriptstyle \pm 0.0026}$
& $0.0065\,{\scriptstyle \pm 0.0193}$
\\

TARNet
& $0.0692\,{\scriptstyle \pm 0.0308}$
& $0.0037\,{\scriptstyle \pm 0.0038}$
& $0.0655\,{\scriptstyle \pm 0.0306}$
& $0.0993\,{\scriptstyle \pm 0.0345}$
& $0.0040\,{\scriptstyle \pm 0.0053}$
& $0.0953\,{\scriptstyle \pm 0.0324}$
& $0.0239\,{\scriptstyle \pm 0.0142}$
& $0.0016\,{\scriptstyle \pm 0.0022}$
& $0.0223\,{\scriptstyle \pm 0.0137}$
\\

GRF / Causal Forest
& $0.0159\,{\scriptstyle \pm 0.0073}$
& $0.0024\,{\scriptstyle \pm 0.0033}$
& $0.0135\,{\scriptstyle \pm 0.0059}$
& $0.0110\,{\scriptstyle \pm 0.0082}$
& $0.0024\,{\scriptstyle \pm 0.0034}$
& $0.0086\,{\scriptstyle \pm 0.0054}$
& $0.0231\,{\scriptstyle \pm 0.0078}$
& $0.0014\,{\scriptstyle \pm 0.0024}$
& $0.0217\,{\scriptstyle \pm 0.0068}$
\\

CFRNet
& $0.0660\,{\scriptstyle \pm 0.0429}$
& $0.0030\,{\scriptstyle \pm 0.0038}$
& $0.0630\,{\scriptstyle \pm 0.0420}$
& $0.1094\,{\scriptstyle \pm 0.0645}$
& $0.0044\,{\scriptstyle \pm 0.0100}$
& $0.1049\,{\scriptstyle \pm 0.0630}$
& $0.0167\,{\scriptstyle \pm 0.0134}$
& $0.0012\,{\scriptstyle \pm 0.0015}$
& $0.0155\,{\scriptstyle \pm 0.0127}$
\\

TabPFN-v3.5 + X-Learner
& $0.0232\,{\scriptstyle \pm 0.0141}$
& $0.0025\,{\scriptstyle \pm 0.0034}$
& $0.0207\,{\scriptstyle \pm 0.0131}$
& $0.0237\,{\scriptstyle \pm 0.0139}$
& $0.0030\,{\scriptstyle \pm 0.0036}$
& $0.0207\,{\scriptstyle \pm 0.0117}$
& $0.0213\,{\scriptstyle \pm 0.0117}$
& $0.0011\,{\scriptstyle \pm 0.0017}$
& $0.0202\,{\scriptstyle \pm 0.0113}$
\\

TabPFN-v3.5 + S-Learner
& \textcolor{bestred}{$0.0008\,{\scriptstyle \pm 0.0010}$}
& \textcolor{bestred}{$0.0005\,{\scriptstyle \pm 0.0008}$}
& \textcolor{bestred}{$0.0003\,{\scriptstyle \pm 0.0004}$}
& \textcolor{bestred}{$0.0015\,{\scriptstyle \pm 0.0009}$}
& \textcolor{bestred}{$0.0005\,{\scriptstyle \pm 0.0005}$}
& \textcolor{secondorange}{$0.0010\,{\scriptstyle \pm 0.0005}$}
& $0.0062\,{\scriptstyle \pm 0.0070}$
& \textcolor{bestred}{$0.0004\,{\scriptstyle \pm 0.0009}$}
& $0.0058\,{\scriptstyle \pm 0.0065}$
\\

TabPFN-v3.5 + DR-Learner
& $0.0159\,{\scriptstyle \pm 0.0133}$
& $0.0020\,{\scriptstyle \pm 0.0027}$
& $0.0139\,{\scriptstyle \pm 0.0126}$
& $0.0247\,{\scriptstyle \pm 0.0170}$
& $0.0024\,{\scriptstyle \pm 0.0030}$
& $0.0223\,{\scriptstyle \pm 0.0157}$
& $0.0153\,{\scriptstyle \pm 0.0109}$
& \textcolor{secondorange}{$0.0010\,{\scriptstyle \pm 0.0017}$}
& $0.0143\,{\scriptstyle \pm 0.0103}$
\\

\bottomrule
\end{tabular}
}
\end{table*}

\paragraph{RQ3: What happens when confounders are hidden?}
\label{sec:rq_hidden}
As illustrated in Figure~\ref{fig:hidden_confounder_track},  synthetic and IHDP comparisons reveal how estimation error changes when a confounder is omitted. Most models show pronounced error increases. Separately, the binary partial-identification companion evaluates recovery of identified-set boundaries. TabPFN-v3.5-based models achieve the lowest endpoint RMSE in the Manski, IV and sensitivity bound settings. Appendix~\ref{app:hidden_confounding} defines the identified sets and Appendix~\ref{sec:hidden} reports the detailed results.

\paragraph{RQ4: Are estimator rankings consistent across semi-synthetic and real-world data?}
\label{sec:rq_external}
Tables~\ref{tab:semisynthetic_backdoor} and~\ref{tab:null_cate_rct}
evaluate CATE accuracy on semi-synthetic data and null-effect
behavior on real-world pre-treatment outcomes, respectively.
Both evaluations show dataset-dependent performance, with no
consistently best estimator.
By mean $\mathrm{sPEHE}^2$, the best TabPFN-v3.5 meta-learner is
the X-Learner on ACIC 2016, the DR-Learner on both LaLonde
benchmarks, and the S-Learner on all three real-world null-effect tests.
Among all evaluated methods, TabPFN-based estimators attain the
lowest error on ACIC 2016, ACTG 175, and Pennsylvania,
whereas CFRNet, CausalPFN, and CausalFM lead on LaLonde-PSID,
LaLonde-CPS, and Illinois, respectively.
The TabPFN-based DR-Learner ranks second overall on both
LaLonde benchmarks. These results support modular estimation as a competitive
alternative to CFMs, while highlighting that the choice of causal
estimator remains consequential even with the predictive backbone fixed.

%% file: sec/7.Conclusion.tex
\section{Conclusion}

We introduce \textsc{CausalIDView} to evaluate how causal estimators
respond to changes in identifying information.
By holding the realized SCM, query units, and target CATE fixed
across observational views, the benchmark enables controlled
comparisons that are difficult to obtain from separately
constructed datasets. Changes in the rankings of the evaluated CFMs highlight the value
of these matched comparisons across identification regimes.
The evaluation also reveals that TabPFN-based estimators achieve
the best point-estimation performance in the back-door,
front-door, and proximal views. These matched comparisons further reveal model-specific failure
patterns as the available identifying information changes.
Such findings motivate CFMs that better account for the
identifying roles of observed variables. Across the semi-synthetic and real-world evaluations,
no single model consistently performs best, highlighting the
need for more reliable generalization across diverse
data-generating settings. Future work should extend \textsc{CausalIDView} to broader causal
settings and further investigate TabPFN-based estimator designs
across identification regimes.